\documentclass{article}
\usepackage{ijcai21}

\RequirePackage[T1]{fontenc} 
\RequirePackage[tt=false, type1=true]{libertine} 
\RequirePackage[varqu]{zi4} 

\RequirePackage[libertine]{newtxmath}
\usepackage[letterspace=-25]{microtype}

\usepackage{graphicx}
\graphicspath{ {./images/} }

\usepackage{savesym}
\savesymbol{Bbbk}

\usepackage{amssymb}
\usepackage{color}
\usepackage{url}
\usepackage{multirow}
\usepackage{xcolor}
\usepackage{dsfont}
\usepackage{algorithm}
\usepackage{algorithmic}
\usepackage{amsmath}
\usepackage{bm}
\usepackage{booktabs}
\usepackage{multirow}
\usepackage{enumitem}

\restoresymbol{NEW}{Bbbk}

\DeclareMathOperator*{\argmax}{arg\,max}

\newcommand{\FTkdd}[1]{\textcolor{black}{#1}}

\newcommand{\FTx}[1]{\textcolor{black}{#1}} 
 
\newcommand{\hidebig}[1]{%#1
}

\newcommand{\AR}[1]{\textcolor{black}{#1}}

\definecolor{cadmiumgreen}{rgb}{0.0, 0.70, 0.25}
\newcommand{\EABC}[1]{\textcolor{black}{#1}}

\newcommand{\todo}[1]{\textcolor{black}{#1}}
\newcommand{\del}[1]{}
\newcommand{\delete}[1]{}
\newcommand{\nos}[1]{}

\newcommand{\ARR}[1]{{#1}}

\def\resp{resp.}

\def\DAX{DAX}
\def\depth{deep-fidelity}
\def\Depth{Deep-fidelity}

\def\feedforward{feed-forward}
\def\Feedforward{Feed-forward}

\newcommand{\Args}{\ensuremath{\mathcal{A}}}
\newcommand{\Influences}{\ensuremath{\mathcal{I}}}

\newcommand{\SF}{\ensuremath{\sigma}}

\newcommand{\Rels}{\ensuremath{\mathcal{R}}}
\newcommand{\Atts}{\ensuremath{\Rels^-}}
\newcommand{\Supps}{\ensuremath{\Rels^+}}
\newcommand{\pSupps}{\ensuremath{\Rels^{+}}}
\newcommand{\arga}{\ensuremath{\alpha}}
\newcommand{\argb}{\ensuremath{\beta}}

\def\user{\chi}

\newcommand{\f}{\ensuremath{f}}
\newcommand{\x}{\ensuremath{x}}
\newcommand{\out}{\ensuremath{o}}

\newcommand{\inff}{\ensuremath{\mathcal{I}}}

\newcommand{\rc}{\ensuremath{c}}

\newcommand{\NN}{\ensuremath{\mathcal{N}}}
\newcommand{\neurons}{\ensuremath{V}}
\newcommand{\links}{\ensuremath{E}}
\newcommand{\sneurons}{\ensuremath{N}}

\def\rep{\rho} %mapping args to neurons
\newtheorem{definition}{Definition}

\usepackage{fancyhdr}
\title{Deep Argumentative Explanations}

\author {
        Emanuele Albini, Piyawat Lertvittayakumjorn, Antonio Rago \& Francesca Toni \\
\affiliations\large
    Department of Computing, Imperial College London, UK\\
\emails
    \{ emanuele, pl1515, a.rago, ft \}@imperial.ac.uk
}

\begin{document}
\maketitle

\begin{abstract}
	
	Despite the recent, widespread focus on eXplainable AI (XAI), explanations computed by XAI methods
tend to provide little insight into the %actual 
functioning of  Neural Networks (NNs). 
We propose a 
novel  framework for 
obtaining (local) explanations from NNs while providing transparency about their inner workings,
and show how to deploy %the framework 
it
for various neural architectures and tasks. We refer to our novel explanations collectively as
 \emph{Deep Argumentative eXplanations} (\DAX s in short), given that they reflect the \emph{deep} structure of the underlying NNs and that they are defined in terms of notions from \emph{computational argumentation}, a  form of symbolic AI offering useful reasoning abstractions for explanation. 
We  
evaluate \DAX s empirically showing that they exhibit   
%along several dimensions for a number of tasks and datasets
 %(notably %importance of their depth
 \emph{\depth} %, stability, 
 and low computational cost. 
We also conduct human experiments %with  \DAX s obtained from convolutional NN models for text classification, 
indicating that \DAX s are comprehensible to humans and align with their judgement, while also being competitive, in terms of user acceptance,  with some existing  
approaches to XAI that also have an argumentative spirit.
	
\end{abstract}

\section{Introduction}

Several  recent efforts in AI are being %dedicated towards 
devoted to explainability, in particular of 
black-box methods such as deep learning. However, the majority of explanation methods in the literature (e.g., see overview in \cite{Guidotti}) focus on describing the role of inputs (e.g., features) towards outputs. A consequence of such ``flat” structure of explanations is very little transparency on how the outputs are obtained.   
This focus on the ``flat'', functional, input-output behaviour also disregards the human perspective on explanations and the potential benefits for XAI from taking a social science viewpoint  \cite{Miller:19}.

To address these %two 
issues when explaining predictions by \emph{Neural Networks (NNs)},
we propose
\emph{Deep Argumentative eX%Ex
planations (\DAX s)},  based on concepts drawn %drawing inspiration 
from the field of Computational Argumentation in symbolic AI (see overview in \cite{AImag}).

%\delete{focusing on their use} to explain Neural Networks (NNs) of various kinds and tasks/data, %\del{but}  all trained for prediction. 
Argumentative models of
explanation %have been advocated, including  argumentative ones 
are  advocated  in the social sciences, e.g., in \cite{argExpl}, as suitable to humans. 
%Given the important role of argumentation in human interactions \todo{ ADD MERCIER?}, argumentative explanations have the potential to increase trust in the AI systems they explain.
Indeed, the explanations generated by some existing methods have an argumentative spirit, e.g.,
LIME \cite{Ribeiro_16} and SHAP \cite{Lundberg_17} assign to input features importance scores towards outputs, giving a form of weighted pro/con evidence.
Computational argumentation is particularly suitable for underpinning argumentative explanations. 
It is a mature field with  solid
%strong theoretical and algorithmic 
foundations, offering a variety of 
 %\del{so-called} 
\emph{argumentation frameworks} comprising sets of arguments and \emph{dialectical relations} between them, e.g., of \emph{attack} (cf. con evidence), as in \cite{Dung:95}, %\delete{and, in addition \cite{Cayrol:05} or instead \cite{supportOnly}, of \emph{support}).}
and/or \emph{support} (cf. pro evidence), as in \cite{Cayrol:05}/\cite{supportOnly}, \resp\
In these frameworks, arguments may be chained by dialectical relations forming potentially deep (debate) structures.  In %\delete{several of these} 
some such frameworks, starting from \cite{Dung:95}, anything may amount to an argument, so long as %\delete{as soon as }
it is in dialectical relation with other arguments (e.g., strategies in games can be seen as arguments \cite{Dung:95}, and, in our \DAX s, neurons and groups thereof in NNs may be understood as arguments).    
Further,
in computational argumentation, 
debates (i.e., argumentation frameworks)
can be analysed using  
 so-called semantics, e.g.,  amounting to definitions of \emph{dialectical strength} satisfying desirable \emph{dialectical properties} 
 (such as that attacks against an argument should decrease its  dialectical strength \cite{Baroni_18}). When argumentation frameworks are extracted for the purpose of explaining underlying methods (e.g., recommenders, as in \cite{rec}) %any quantities can serve as semantics 
 \AR{a} semantics is drawn from quantities computed by the underlying methods 
 (e.g., predicted ratings in \cite{rec}) insofar as they satisfy %desirable, 
 relevant  dialectical properties. 

\begin{figure*}[h]
    \centering
    {\includegraphics[width=0.855\textwidth]{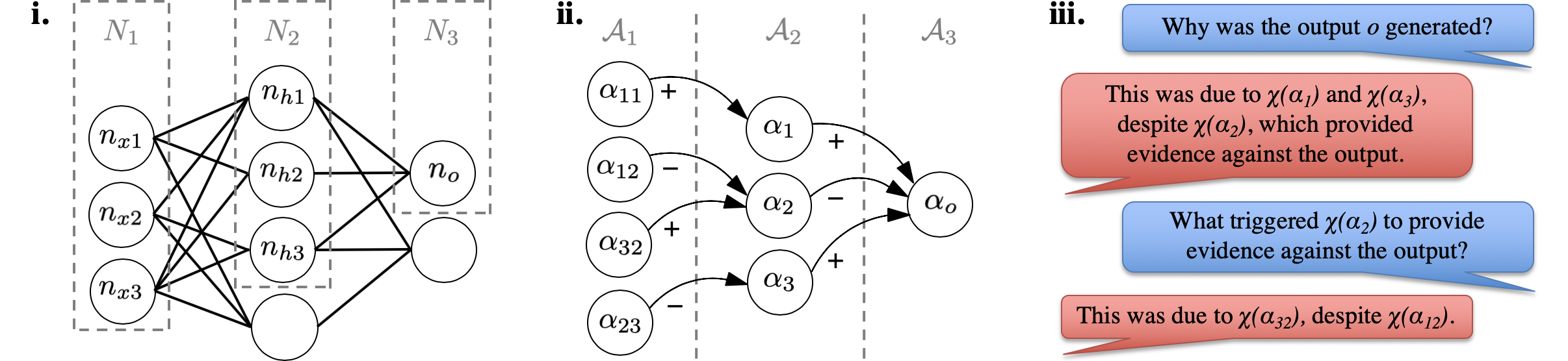}}
    \caption{(i) %Example 
    Toy \feedforward\ NN  (FFNN), with input/hidden/output neurons selected as nodes labelled with subscript $x/h/o$, \resp, and %strata 
    layers $N_1,N_2,N_3$; (ii) extracted Generalised Argumentation Framework (GAF) with dialectical relations of attack (indicated with 
    $-$) and support (indicated with
    $+$); and (iii) example %\delete{conversational} 
    DAX generated from the GAF (with a conversational format% \ARR{$\phi$})
    , comprising user (right) and system (left) statements).
    Here, $\user$ (a hyper-parameter in our framework) indicates a representation,  amenable to user consumption, for arguments in the GAF, dependent on the cognitive abilities and needs of the target users% of \DAX s
    .
    }
    \label{fig:eg}
\end{figure*}

Figure~\ref{fig:eg} gives a simple illustration, for a toy \feedforward\ NN (FFNN), of our framework (see Section~\ref{sec:methodology} for details). 
In a nutshell, \DAX s (in various formats, e.g., conversational as in Figure~\ref{fig:eg}) are extracted from
(instances of) \emph{Generalised Argumentation Frameworks} (GAFs) potentially including any number and type of dialectical relations (e.g., attack and support, as in Figure~\ref{fig:eg}). 
\del{These consists}
\AR{These frameworks consist} of directed links between (groups of) neurons (treated as arguments)  relevant to the NN prediction in needs of explanation. 
GAFs are extracted so that arguments therein are equipped with a  dialectical strength given by
 quantitative  measures derived from the underlying NNs, e.g., in one of the DAX instances we propose, given by {Layer-wise Relevance Propagation} (LRP) \cite{Bach_15}),
 governed by desirable dialectical properties identified up-front to fit the setting of deployment for the given NNs/tasks.
%Note that whereas in Figure~\ref{fig:eg} arguments in the GAF all represent individual neurons and all layers in the FFNN contribute arguments to the GAF, in general arguments may be chosen to represent groups of neurons, possibly from selected layers only. Also, whereas in Figure~\ref{fig:eg} the dialectical relations in the GAF amount to attack and support only, in general GAFs may include any dialectical relations. Further, in Figure~\ref{fig:eg}, \DAX s take a format suitable to empowering conversations with humans, but we will also consider other formats in the paper.

\iffalse
To obtain DAX, our proposed methodology consists of three steps.
First, we identify, in the trained network, (groups of) neurons and connections between them which are relevant to the output to be explained. 
Second, we extract a general argumentation framework (GAF) which is comprised of a set of arguments and sets of dialectical relations, governed by desirable dialectical properties.
Third, we generate DAX, which is amenable to human consumption, from GAF.
\fi

Our contribution in this paper is threefold: (1) we give the general \DAX\ framework and three  concrete  instances thereof\nos{ (all of depth 3, as in Figure~\ref{fig:eg})}, 
%with CNNs for text classification, with CNNs for image classification, and with  FFNNs for prediction with tabular data.
 varying the underlying NN, the GAFs,  how their semantics is derived from the underlying NNs, the dialectical properties it satisfies, and the format in which \DAX s are delivered to users; 
 %In all three settings  we show that  \DAX s are stable \cite{Sokol_20},
 (2)  we evaluate the three instances empirically, showing that they exhibit a property that we term \emph{\depth} 
 %i.e. , every time the same input-output pair needs to be explained (by a fixed trained NN) the same explanation is generated, 
 and that they have low complexity; %in that, after initial set-up, \DAX s can be generated efficiently. 
%\todo{Further, we show that \DAX s' depth is important: when considering samples with ``similar'' inputs and outputs, in most cases the intermediate nodes in the explanations change considerably to reflect changes in the model activation to determine the output (we call this novel property \emph{\depth})}. \todo{we probably will need to refine the \depth\ experiments...}
% 
(3)
%in the text classification setting, 
we conduct experiments with humans showing some of \DAX%’s
s' advantages: comprehensibility to humans and alignment with their judgement, as well as, 
in comparison with
LIME  and SHAP (selected \nos{for comparison} due to their argumentative spirit), their increased
ease of understanding, insightfulness, and capability of inspiring trust\nos{ and potential for visual appeal}. 

Although several works exist for argumentation-based explanation% in the literature
, e.g., \cite{Briguez:14,rec}, to the best of our knowledge \DAX s are the first for NNs.
We plan to release the \DAX\ toolkit upon acceptance.

\section{Background and Related Work}  
\label{sec:background}

\hspace*{0.3cm} \emph{%(Artificial) 
Neural Networks (NNs).}
\label{background:NN}
For our purposes a (trained) NN \NN\
can be seen  as a directed graph
 whose nodes are neurons and whose edges are connections between neurons, formally \NN\ is  
$\langle \neurons,\links \rangle$ %consisting of 
 with \neurons\ a set of  neurons and $\links\subseteq \neurons \times \neurons$ a set of directed edges% between neurons
 . 
 Different neural architectures impose different restrictions on these graphs (e.g., in the FFNN in Figure~\ref{fig:eg}i, neurons are organised in strata with each neuron connected to every neuron in the next stratum).   
  Independently of the architecture, neurons in a trained NN are equipped with activation values 
  (computed by some activation function) and  links are  equipped with weights (learnt during training).\footnote{Note that, given that we focus on explaining trained NNs, we ignore here how training is performed, and how performant the \emph{trained} NNs are. Also, for simplicity we ignore biases %in the trained NNs 
  in the formalisations in the paper.}

\emph{Computational argumentation.}
\label{background:arg}
\DAX s are %defined in terms of
obtained from
Generalised Argumentation Frameworks (GAFs),
e.g., 
as understood in \cite{Gabbay:16%\nos{,Baroni_17
}. Formally, these are 
tuples $\langle \Args, \Rels_1, \ldots, \Rels_m \rangle$  with $\Args$ a set (of \emph{arguments}), $m \geq 1$ and, $\forall i \in \{1, \ldots, m\}$, 
$\Rels_i \subseteq  \Args \times \Args$ is a binary, directed \emph{dialectical relation}.
Intuitively, GAFs are abstractions of debates, with arguments representing opinions and dialectical relations expressing various forms of agreement (support) or disagreement (attack) between opinions. GAFs can also be seen as directed graphs (as in Figure~\ref{fig:eg}ii) , with arguments as nodes and dialectical relations as labelled edges.  
Various instances of GAFs have been studied for specific choices of dialectical relations.
A well-studied instance is \emph{abstract argumentation frameworks} \cite{Dung:95} with $m=1$ and $\Rels_1=\Atts$ \AR{is} attack.
We will use 
%\emph{abstract argumentation frameworks} \cite{Dung:95} with $m=1$ and $\Rels_1$ attack,  
\emph{support argumentation frameworks} (SAFs)~\cite{supportOnly}, where $m\!=\!1$ and $\Rels_1=\Supps$ is support; 
\emph{bipolar argumentation frameworks} (BAFs) \cite{Cayrol:05}, where $m\!=\!2$, $\Rels_1=\Atts$ is attack,  $\Rels_2=\Supps$ is support;
and a form of \emph{tripolar argumentation frameworks}% (TAFs) \cite{rec} 
, where $m\!=\!3$\nos{, $\Rels_1=\Atts$ is attack, $\Rels_2=\Supps$ is support, and $\Rels_3=\Rels^!$ is a form of support that we call \emph{critical support}}.
We will also use %notions 
%the notion of \emph{gradual semantics}, 
semantics given by mappings $\SF: \! \Args \! \rightarrow \! \mathcal{V}$ for evaluating arguments' \emph{dialectical strength} over a %\nos{(possibly infinite)}
set of values $\mathcal{V}$ \cite{Baroni_18}, (in this paper $\mathcal{V}=\mathbb{R}$ and \SF\ is extracted from trained NNs). Finally, we will use (existing and novel) \emph{dialectical properties}, in the spirit of \cite{Baroni_18}, that \SF\ needs to satisfy to ensure that the GAFs have natural features of human debates (e.g., that strong attacks decrease the strength of opinions). 

\begin{figure*}[t]
    \centering
    {\includegraphics[width=0.85\textwidth]{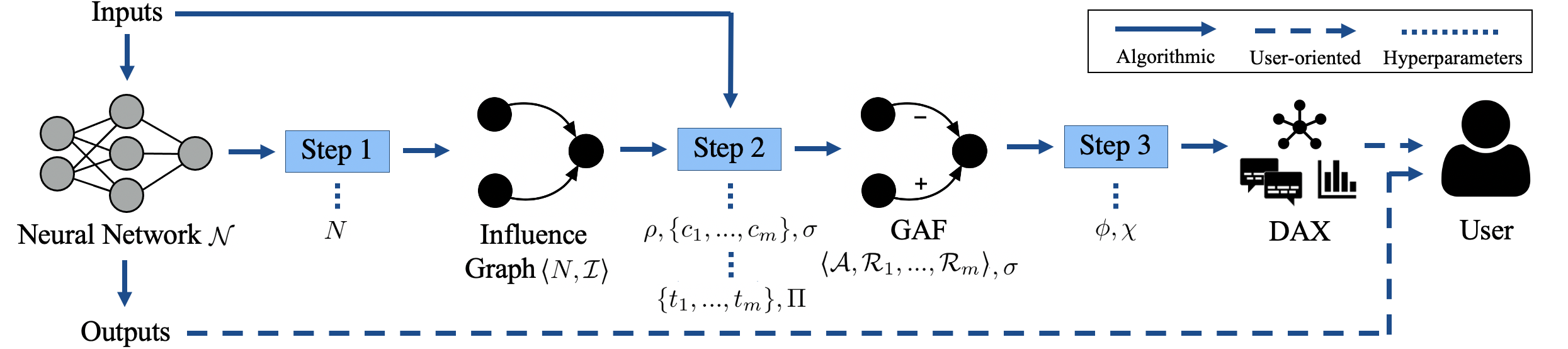}}
    \caption{\DAX\ %methodology
    framework (alongside the typical process of obtaining outputs from a neural model given its inputs) comprising steps:
    1. 
% Identify, within \NN, \emph{nodes} and \emph{influences} between them
Based on the chosen \emph{nodes} $N$ within \NN, extract $\langle N, \inff \rangle$, a directed graph of \emph{influences} between nodes%\delete{, as a pool from which to draw explanations}
.
2. Extract a \emph{GAF} from the output of the first step, 
based on choices of \emph{argument mapping} $\rho$, \emph{relation characterisations} $\{c_1,...,c_m\}$ and \emph{dialectical strength} $\sigma$. These choices are driven by the types of relations $\{t_1,...,t_m\}$ to be extracted and the \emph{dialectical properties} $\Pi$ that %\delete{the GAF} 
$\sigma$ should satisfy (on the GAF) to form the basis for explanations. 
3. Generate a \emph{\DAX} from the GAF and dialectical strength $\SF$, for user consumption in a certain \emph{format} $\phi$, associating arguments with human-interpretable concepts through a \emph{mapping} $\chi$. }
    \label{fig:dig}
\end{figure*}

\emph{XAI approaches.}
Amongst the many proposals  (e.g., see survey in \cite{Guidotti}), some are model-agnostic (e.g., LIME \cite{Ribeiro_16}, SHAP \cite{Lundberg_17}\nos{, CXPlain \cite{Schwab_19}}), while others are tailored to specific methods (e.g., for NNs, %\delete{Layer-wise Relevance Propagation (LRP)} 
LRP \cite{Bach_15}, \nos{contrastive explanations \cite{Dhurandhar_18}, and} Grad-CAM \cite{Selvaraju2016}\FTkdd{, GRACE~\cite{kdd2020}}),
%\delete{The explanations generated by some of these methods have an argumentative %\del{flavour} 
%\AR{spirit}, e.g.\FTa{,}
%LIME and SHAP assign to input features importance scores towards outputs, giving a form of weighted pro and con evidence, and contrastive explanations %\cite{Dhurandhar_18} 
%use pertinent positive and negative features}. 
%All these methods 
but all generate ``flat'' explanations,  in terms of  input features only, without %any indication of 
indicating
how. %they are generated. 

We will compare DAX\ARR{s} with %both 
LIME and SHAP in %some of 
our human experiments, and will use LRP and Grad-CAM to compute dialectical strength in some of our proposed DAX instantiations. 
In a nutshell, LIME and SHAP compute (using different methods) positive and negative attribution values for input features, when explaining outputs by any classifier (the details of these computations are not important here). For both methods, the computed (positive and negative) attribution values can be seen as arguments (for and against, \resp, the output prediction). This argumentative spirit of LIME and SHAP is the reason why we choose to compare them with \DAX s in the human evaluation.
\EABC{LRP is a technique for interpretability \AR{which is able to scale }\del{scaling }to \del{potentially }highly complex NNs: it operates by propagating \AR{a }\del{the }prediction backward\AR{s} in the NN, using a set of purposely designed propagation rules (specifically we use LRP-0 \cite{Bach_15}\nos{, see details in Appendix A}). %
 Grad-CAM %(Gradient-weighted Class Activation Mapping)} 
 uses the gradients of any target concept (e.g., ``dog''% in a classification network or a sequence of words in captioning network
 ) flowing into the final convolutional layer of a CNN {for image classification} to produce a coarse localisation map highlighting the important regions in the input image for predicting the concept\nos{ (see details in Appendix A)}.}

Some explanation methods, like \DAX s, make use of symbolic representations, e.g., \nos{influences between variables in \cite{BC},}  ordered decision diagrams in \cite{Shih_18} %for Bayesian network classifiers 
and logical representations in \cite{Ignatiev_19}% and \todo{...decision trees}
. These methods also focus on input-outputs only and are ``flat''.  %\del{mention we take approach of ML as decision functions as they do???}
Computational argumentation is used by many for explanation purposes (e.g.,  %recently 
in \cite{Naveed_18%nos{,rec,Cyras_19,Madumal_19
}). 
Various neural architectures  have also been used for argument mining from text 
\cite{AM-TorroniLippi%nos{,AM,AM-Reed
}, in some cases to extract GAFs \cite{AM-Oana}, but not for the purposes of explanation.
%(but from %textual data 
%text exclusively).
%However, 
%{Overall}, to the best of our knowledge, argumentation has not been used \ftai{so far} to explain the outputs of NNs.

Some recent methods generate ``structured'' explanations, including hierarchical explanations~\cite{Chen_20X} and explanations exploiting dependencies \cite{dependencies}. These incorporate interactions between input features only%, rather than from hidden layers
, so they can also be deemed ``flat'' in our sense.

%\del{ \emph{Hierarchical explanations} \cite{Chen_20} have also been defined for various neural classifiers, where interactions between features are detected. Clearly, these methods vary in the information they include in explanations, but this information is still limited to parts of the inputs. 
%\todo{FT: add also mention of other works trying to get explanations with "dependencies", e.g.https://arxiv.org/pdf/1903.10464.pdf
%https://arxiv.org/pdf/1910.06358.pdf}}

Some works, like \DAX s, move away from  ``flat'' explanations, predominantly with images: in \cite{Wang_19_NIPS} regions of an image generating insecurities between two outputs are highlighted, thus unearthing some of the ``deliberative'' reasoning behind the outputs (but still focusing on  input features); network dissection \cite{Bau_17} identifies regions of an image activating certain filters in a Convolutional NN (CNN), giving these filters a semantic meaning; {self-explaining NNs} \cite{Alvarez-Melis_18} use one NN for classification and another for explanation, to obtain explanations based on higher-level features/concepts and their relevance towards a classification; in \cite{Olah_18} positive/negative connections between %these 
%similar 
concepts are extracted from neurons in a CNN%YES \todo{is this about images too?}
. 
%\todo{ADD https://arxiv.org/pdf/1901.02413.pdf?}
There are promising steps, but focus on visualisation, falling short of a general %methodology 
framework for explanations showing the interplay amongst inputs, intermediate layers and outputs.

%\todo{Should we also mention early work on trace-based explanations in AI? e.g. for expert systems (such ashttps://www.sciencedirect.com/science/article/abs/pii/0004370283900085 ?)}}

%ALSO from ACL2020 ??? https://arxiv.org/pdf/2004.02015.pdf

Several works  advocate the need for evaluating explanations (e.g., see overview in \cite{Sokol_20}) %, alongside several dimensions
in several ways. We consider \AR{the }\nos{stability and }computational cost (of obtaining \DAX s from trained NNs), \nos{both }\del{standardly }\AR{commonly }used in the literature \cite{Sokol_20}, as well as a novel property of \emph{\depth}%\ for deep explanations
, evaluating it empirically for our proposed \DAX\ instances.

A user-driven perspective on explanations is advocated by many \cite{Miller:19}, as is the fact that different explanation styles, as envisaged by \DAX, may need to be considered to aid transparency and trust and to support human decision making (e.g. see \cite{Rader_18,%piyawat,
Kunkel_19}).
In the same spirit, we conduct a human study on a \DAX\ instance.

\section{
\DAX\  %General 
Framework}

\label{sec:methodology}
%%%%%%%%%%%%%%%%%%%%%%%%%% 

% FOR FIGURE
\iffalse
$| \quad\quad\quad\quad\quad\quad\quad\quad\quad\quad\quad\quad  |$ \\ \\
$\mathcal{N}$ \\ \\
$N$ \\ \\
$\langle N, \mathcal{I} \rangle$ \\ \\
$\rho, \{ c_1, ... , c_m \}, \sigma$ \\ \\
$\{ t_1, ... , t_m \}, \Pi$ \\ \\
$\langle \Args, \mathcal{R}_1, ..., \mathcal{R}_m \rangle$ \\ \\
$\phi, \chi$ \\ \\
\fi

Here, we %define 
outline our general %methodology 
\DAX\ framework 
for explaining NN predictions. 
Throughout, we use the \nos{simple} example in Figure~\ref{fig:eg} for illustration, and otherwise
assume as given a generic, trained NN \NN\ for prediction%\delete{, of whichever kind and for whichever task/data}
, with explanations  needed for an 
output \out\ computed by %the given NN 
\NN\ for \nos{some} given input \x, where \out\ is a single classification, e.g., the most probable class in the output layer of \NN.
For the purpose of explaining \out\ for \x, \NN\ amounts to a function $\f$ such that $\f(\x) = \out$ and can be equated to a  directed graph $\langle \neurons,\links \rangle$ (as described in Section~\ref{background:NN}).

\DAX s are constructed in 3 steps, as %indicated 
outlined in Figure \ref{fig:dig} and described below, with a focus on the choices (hyper-parameters) that need to be instantiated when \DAX s are deployed.

\emph{Step 1: from NNs to influence graphs.}
%
%Step 1
 This
 amounts to %selecting
 determining, within \NN, an \emph{influence graph} $\langle \sneurons, \Influences \rangle$ with a set  $\sneurons$ of nodes and a set  $\Influences$ of influences as candidate elements of (binary) dialectical relations of the GAF (to be identified at Step 2) underlying  explanations for why $\f(\x)=\out$. 
 The choice of  $\sneurons$  %WRONG \subseteq
 within
 $\neurons$ is a hyper-parameter dictated by the setting of deployment. 
 We will see that in some instantiations of \DAX\ $\sneurons \subseteq \neurons$, %these nodes are neurons; 
 whereas
 in others nodes in $\sneurons$ may also include groups of neurons in $\neurons$. Note that,  in principle, $\sneurons$ may be exactly $\neurons$  but in practice it will need to be considerably smaller, to accommodate the cognitive needs of users.
  Once \sneurons\ has been chosen, we obtain \Influences\ automatically %from paths in $\langle \neurons,\links \rangle$ between (neurons in) nodes. Formally:
  as follows:

\begin{definition}
\label{def:influences}
 Let hyper-parameter $\sneurons \subseteq \neurons \cup \mathcal{P}(\neurons)$ be given. Then, the \emph{influence graph} $\langle \sneurons, \Influences \rangle$ is such that  
 \Influences\ =
%We define the set \inff\ of influences %\ARR{, and thus the \emph{influence graph} $\langle N, \Influences \rangle$,} to be  
$\{(n_1,n_2) | n_1,n_2 \in \sneurons$ and there is a path in $\langle \neurons,\links \rangle$ from (a neuron in) $n_1$ to (a neuron in) $n_2$%
%IN FULL either $n_1,n_2 \in \neurons$ and there is a path%\footnote{\todo{shall we add definition of path?}} in $(\neurons,\links)$ from $n_1$ to  $n_2$, or $n_1 \in \neurons$  and $n_2 \subseteq \neurons$ and there is a path in $(\neurons,\links)$ from  $n_1$ to  some $n_2' \in n_2$, or $n_1 \subseteq \neurons$  and $n_2 \in \neurons$ and there is a path in $(\neurons,\links)$ from some $n_1'\in n_1$ to  $n_2$, or $n_1 \subseteq \neurons$  and $n_2 \subseteq \neurons$ and there is a path in $(\neurons,\links)$ from  some $n_1' \in n_1$ to  some $n_2' \in n_2$
$\}$.
\end{definition}

 Practically, $\sneurons$ will include (groups of) neurons in the input layer of \NN,  the neuron $n_{\out}$ in the output layer responsible for output \out\  as well as any  %\todo{``semantically meaningful'' AR BETTER TERM?} 
 \AR{desired} %\todo{AVOIDS THE ISSUE?} 
 (groups of) neurons from the hidden layers. 
 For example, if \NN\ is a CNN, then $\sneurons$ may consist of 
 the neurons in the input layer, $n_\out$ as well as all the the neurons of the max-pooling layer (as for the  \DAX\ instances \FTkdd{for text, in Section~\ref{CNN-text}} % for text classification we will define
 ) or all filters in the last convolutional layer (as for the  \DAX\ instance %for image classification we will define
 \FTkdd{for images, in Section~\ref{CNN-image}}). Both choices (as well as several other \FTkdd{``semantically meaningful''} % BETTER TERM? \todo{KEEP IT HERE?} 
 alternatives) reflect the inner structure of the underlying NN.
 %\footnote{For simplicity of presentation, in this paper we ignore bias. ...NOT REALLY, if you want bias to be included just consider it as a node, from which an argument can be drawn} 

%~\ref{CNN-text}, \ref{CNN-image} and \ref{FF} \todo{To be changed if we change the empirical}
%\FT{Section~\ref{sec:instances}}).
%

In all \DAX\ instances in this paper, we choose $\sneurons=\sneurons_1\cup \ldots \cup \sneurons_k$, for $k>2$, with $\sneurons_k = \{n_\out\}$ and  $\sneurons_1, \ldots, \sneurons_k$  disjoint sets of nodes %\nos{(i.e. $\sneurons_i\cap \sneurons_j = \emptyset$, for all $i\neq j = 1, \ldots, k$)} 
such that for every $(n_1, n_2) \in \inff$ there exists $i\in \{1, \ldots, k-1\}$ with $n_1 \in \sneurons_i$, 
$n_2 \in \sneurons_{i+1}$. Thus, $\sneurons_1, \ldots, \sneurons_k$ amount to \emph{strata}, contributing (in Steps 2 and 3) to  \emph{deep} explanations, where (arguments representing) nodes in each stratum are explained in terms of (arguments representing) nodes in the previous stratum.
%{Note also that, when using strata, we are basically understanding $\f$ as the composition of $k-1$ functions ($\f = \f_{k-1} \circ ... \circ \f_1$).}
%
%\begin{example}[Hyper-parameter $\sneurons$]
For illustration, 
given the %simple %\feedforward\ 
FFNN in Figure \ref{fig:eg}i, \sneurons\ amounts to all neurons in $\neurons$ with a path to $n_\out$, arranged in three strata to match the three layers of the FFNN. 
Note that, in general, 
depending on the choice of $\sneurons$, strata may result from non-adjacent layers in \NN. 
Also, although the choice of \sneurons\ (and thus \inff) is tailored to explaining $\f(\x)=\out$,  differently from our %simple
earlier illustration the choices %of candidate nodes in \sneurons\ 
from the hidden layers of the given \NN\  can be made a-priori, independently of any input-output pair, and then instantiated for each specific input-output pair (this is exactly what we do in the \DAX\ instances in Section~\ref{sec:instances}, to guarantee a low computational cost, as discussed in Section~\ref{sec:empirical}).

\emph{Step 2: from influence graphs to (dialectical property-compliant) GAFs.}
%Step 2 
This
amounts to extracting a GAF $\langle \Args, \Rels_1, \ldots, \Rels_m \rangle$ from $\langle \sneurons, \inff \rangle$, with 
%\EAB{two} simultaneous choices% driving the extraction of the GAF $\langle \Args, \Rels_1, \ldots, \Rels_m \rangle$ from 
%\EAB{the \emph{\ARR{influence graph}% of influences
%} 
%$\langle \sneurons, \inff \rangle$}\del{strata $\sneurons_1, \ldots, \sneurons_k$}.
% for a specified neuron to be explained $n_{exp} \in \sneurons_k$
%One choice concerns arguments and dialectical relations: the former are 
arguments drawn from the nodes in $\sneurons$ and %the latter 
dialectical relations drawn from the influences in \inff%\ (for a simple example see Figure \ref{fig:eg}ii).
. This requires, first and foremost, the choice (hyper-parameter) of the number of dialectical relations ($m$) and their   \emph{relation types} $t_1, \ldots, t_m$. 
%
%\begin{example}[Hyper-parameters $t_1, \ldots, t_m$]
For illustration, 
In Figure~\ref{fig:eg}ii the GAF is a BAF with two dialectical relations with relation types attack (-) and support (+).
The choice of the types indicates that the envisaged \DAX s will be defined in terms of (dis)agreement   between  nodes (i.e., neurons or groups thereof).
%\end{example}

%We drive the 
The extraction of arguments is driven by %means of 
a (hyper-parameter) mapping  $\rep: %\del{\Args_1 \cup \ldots \cup \Args_k}
\Args \rightarrow 
%\del{\sneurons_1  \cup \ldots \cup \sneurons_k}
\sneurons$ such that $\Args = \Args_1 \cup \ldots \cup \Args_k$ and 
for each $\alpha \in \Args_i$ there is exactly one $n \in \sneurons_i$ with $\rep(\alpha)=n$ (we say that  %{$n$ corresponds to $\alpha$, and that
$\alpha$ \emph{represents} $n$ and that $\Args_i$ \emph{represents} $\sneurons_i$). 
%Intuitively, arguments represent nodes and are partitioned so as to mirror the strata in the influence graph.
%
%\begin{example}[Hyper-parameter $\rep$]
%\label{ex:rho}
For illustration, in Figure~\ref{fig:eg}ii 
all nodes in \sneurons\ contribute at least one argument, with $n_{\x 1}$ contributing two arguments, and all other nodes contributing exactly one argument. Thus, $\rep(\alpha_{ij})=n_{\x i}$, $\rep(\alpha_i)=n_{hi}$, for $i=1,2,3$, and $\rep(\alpha_\out)=n_\out$. 
%\end{example}

%We %formalise the requirements for a dialectical relation 
%drive the 
The extraction of $\Rels_{j}$
is driven by %means of 
a (hyper-parameter) \emph{relation characterisation} $\rc_{j}: \Influences \rightarrow \{ true, false \}$ for \emph{relation type} $ t_j$, with $j \in \{1, \ldots, m \}$. 
%
\iffalse
BETTER WHEN WE GIVE INSTANCES...
Possible definitions of dialectical relations of types support (+) and attack (-), as required in our running example, may be as follows
\todo{
\begin{definition}
activition-based...
\\
LRP-based...
\end{definition}}
\fi
For illustration, 
%\begin{example}[Hyper-parameters $\rc_1, \ldots, \rc_m$] 
in Figure \ref{fig:eg}ii, dialectical relations of types support (+) and attack (-) may be defined by  relation characterisations $\rc_+$ and $\rc_{-}$ such that:\footnote{With an abuse of notation, \nos{for simplicity} we write $\rc_{j}(a,b)$ instead of $\rc_{j}((a,b))$\nos{ throughout}, for any $j$.}
\begin{itemize}
\item 
$\rc_+(n_1,n_2)$ is $true$  iff 
$(n_1,n_2) \in \inff$ and, for $a_1$ given by $n_{1}$'s activation and $w_{12}$ the connection weight between $n_1$ and $n_2$ in \NN,  
it holds that $w_{12} a_1 >0$; 
\item $\rc_-(n_1,n_2)$ is $true$  iff 
$(n_1,n_2) \in \inff$ and, for $a_1$ and $w_{12}$ as earlier,  
it holds that $w_{12} a_1 <0$.
\end{itemize}
Thus, influences between nodes become dialectical relations of support (or attack) between the arguments representing them if the product of the activation of the ``influencing'' node and the weight of the connection with the ``influenced'' node (in the trained, underlying \NN) is positive (or negative, \resp). %\todo{for this to be self-contained we need to give these activations and nodes for Figure~\ref{fig:eg}???}
%\todo{DO WE NEED TO ALSO GIVE ACTIVATIONS IN THE RUNNING EXAMPLE FOR THINGS TO MAKE SENSE? ELSE HOW CAN THE READER SEE THAT THE PROPERTY IS SATISFIED?} \AR{I think it's okay without, we would have to add weights too I guess? we could mention a specific activation function which means this is satisfied though?}
%
Then, in the \emph{extracted GAF},   %$\alpha_1$, 
an argument representing node $n_1$ supports %$\alpha_2$,
an argument representing node $n_2$ iff $\rc_+(n_1,n_2)=true$ (similarly for attack and $\rc_-)$.
Note that we
call the  relation characterised by $\rc_+$ `support' (and by $\rc_-$ `attack') as it is defined in terms of agreement (disagreement, \resp) between nodes and thus carries positive (negative, \resp) influence% %, and the second relation, characterised by $\rc_-$, `attack' as it is defined in terms of disagreement and carries negative influence
; the resulting  
relations are thus dialectical. 
%\end{example}

In this paper, 
we require %\footnote{\todo{is this just a restriction for this paper? we envisage...future work...dropping this restriction?}} is 
that %the choices  
%of $\SF$ and $\langle \Args, \Rels_1, \ldots, \Rels_m \rangle$, with accompanying $\rep$ and $ \rc_1, \ldots, \rc_m$, 
the choices allow to extract GAFs which (when seen as graphs with edges of $m$ %different 
kinds) are trees with the %explained node
output's argument at the root (%see Figure \ref{fig:eg}ii for illustration
as %is the case 
in Figure~\ref{fig:eg}ii). 
Formally:

\begin{definition}
\label{def:GAF}
\label{def:args}
Let hyper-parameter\AR{s} $\rep$, $t_1, \ldots, t_m$ and $\rc_1, \ldots, \rc_m$ be given. 
Then \emph{the extracted GAF}  $\langle \Args, \Rels_1, \ldots, \Rels_m \rangle$ is such that:
%(for $\Args = \Args_1 \cup \ldots \cup \Args_k$ where $\Args_i$ is the set of arguments representing the nodes at stratum $\sneurons_i$),  with accompanying $\rep$ and $\rc_1, \ldots, \rc_m$,  need to be such that:
%%%%%%%%%%%%%%%%
    %
    \begin{itemize}
     \item % ALREADY GIVEN BEFORE 
     $\Args = \Args_1 \cup \ldots \cup \Args_k$ where $\Args_i$ is the set of arguments representing the nodes at stratum $\sneurons_i$; % LATER Intuitively, arguments represent nodes and are partitioned so as to mirror the strata.
    \item $\Args_k=\{\alpha_\out\}$ such that $\rep(\alpha_\out)=n_\out$; %LATER Intuitively, the single argument from the last stratum  represents the output node.
    \item $\forall i \in \{2, \ldots, k\}$, $\forall \arga_h \in \Args_i$ and $\forall n_g \in \sneurons_{i-1}$:
    \\
    %\hspace*{0.1cm} 
    $\exists \alpha_g \in \Args_{i-1}$ such that $\rep(\alpha_g) = n_g$ 
    iff 
    \\
    %\hspace*{0.1cm}  
    $\exists (n_g, %\rep(\alpha_h)
    n_h) \in \Influences$% and 
    , $\exists l \in \{ 1, \ldots, m \}$ such that $\rc_l(n_g, %\rep(\alpha_h)
    n_h) = true$% for any $l \in \{ 1, \ldots, m \}$
    %, the latter of which results in $(\alpha_g, \alpha_h) \in \Rels_l$, 
    ; 
    \\
    %\hspace*{0.1cm} 
    also, $(\alpha_g, \alpha_h) \in \Rels_l$, for $n_h=\rep(\alpha_h)$; 
    % \\LATER Intuitively,  %if we begin with the output node in our GAF, as influences from any node in the previous stratum to any node in our framework satisfy any relation characterisations, the node in the previous layer and the influence are represented by an argument with a relation (corresponding to the relation characterisation) in the GAF.
    %nodes in (a stratum in) the influence graph are represented by arguments in the GAF iff they influence nodes  (in the next stratum) represented by arguments themselves (because ultimately they lead to influences towards the output node) that contribute to one of the dialectical relations.
    %%%%%%%%%
    \item $\forall (\alpha_p, \alpha_q), (\alpha_r, \alpha_s) \in \Rels_1, \ldots, \Rels_m$ if $\rep(\alpha_p) = \rep(\alpha_r)$ and 
    
    $\rep(\alpha_q) \neq \rep(\alpha_s)$ then  $\alpha_p \neq \alpha_r$. 
    %LATER \\ Intuitively, a node influencing several nodes may need to correspond to several arguments (one for each influence that makes it into the dialectical relations, e.g., see   Figure~\ref{fig:eg}, with two arguments $\alpha_{11}$ and $\alpha_{12}$ representing $n_{h1}$  to distinguish between its influences on $n_{h1}$ and  $n_{h2}$).
\end{itemize}
\end{definition}
Intuitively arguments represent nodes and are partitioned so as to mirror the strata (first bullet);
the single argument from the last stratum  represents the output node (second bullet); 
nodes in (a stratum in) the influence graph are represented by arguments in the GAF iff they influence nodes (in the next stratum) represented by arguments themselves (because ultimately they lead to influences towards the output node) that contribute to one of the dialectical relations third bullet);
a node influencing several nodes may need to correspond to several arguments, one for each influence that makes it into the dialectical relations (last bullet); for illustration, in Figure~\ref{fig:eg}, two arguments $\alpha_{11}$ and $\alpha_{12}$ represent $n_{\AR{x}1}$ to distinguish between its influences on $n_{h1}$ and $n_{h2}$.

%In particular, this may require that (conceptually) multiple arguments represent the same node (as in the running example). 
Note that, by Definition~\ref{def:GAF},
nodes which are not connected by influences do not contribute to dialectical relations in the GAF and nodes dialectically disconnected from the output node are not represented as arguments in the GAF. This secures \AR{the} ``relevance'' of \AR{arguments in} 
explanations drawn from the GAF.
Note also that, given $\rep$, relation types and characterisations, there is a single extracted GAF according to Definition~\ref{def:GAF}, whose depth reflects the depth of the influence graph from the first step. 
%TRUE BUT NO SPACE, A BIT OF A DETAIL, AND WEAKENS US SOMEWHAT?  \todo{For some (undesirable) choices of relation types and characterisations, some strata in the influence graph may contribute no arguments,   giving rise to ``aborted'' explanations of outputs (not ``going back'' to inputs). Thus, the choices of the hyper-parameters need to guarantee that each $\Args_i$ in $\Args$ is non-empty, for useful \DAX s to be obtained at step 3.}

Step 2 requires two final choices (\AR{of} hyper-parameters), strongly coupled with the choice of relation types and characterisations, and  amounting to  \emph{(dialectical) strength} (for arguments in the extracted GAF) and \emph{dialectical properties} (see Section~\ref{background:arg}) for the chosen strength, guiding and regulating its choice to give rise to dialectically meaningful extracted GAFs.
The notion of strength $\SF: \Args \rightarrow \mathbb{R}$
%, representing 
amounts to a quantitative measure derived from \NN\ and  
giving dialectical meaning to the arguments in \Args\
%the arguments' strength may be the corresponding neurons' activation values
%, then how each of the neuron's affect one another's activations may be represented depending on the  extracted dialectical relations: our next choice
in the context of the chosen dialectical relations.
%
%\begin{example}[Hyper-parameter $\SF$]
%\label{ex-sigma}
In the running example, we can use activations to define $\SF$ too, by setting, for any $\alpha \in \Args$, $\SF(\alpha)=a$  with $a$ the modulus of the activation of $\rep(\alpha)$ in \NN.%
%\end{example}
\footnote{When strata are drawn from non-adjacent layers in \NN\ (see
%as in 
Section~\ref{sec:instances}), %different  
measures of strength/relation characterisations other than using activation\AR{s}  will be needed.}

The choice of dialectical properties $\Pi$ %{for the chosen dialectical strength}, 
determine how ``natural'' the \DAX\ obtained at Step 3 will be.
%%%%%%%%%%%%
\begin{table}[h!]
    \centering
    \begin{tabular}{|l|l|}
     \hline 
     \!\!\!\emph{Dialectical}\!\!\! & $\forall \arga, \argb \in \Args$: \\
     \!\!\!\emph{Monotonicity}\!\!\! &
     %Then, \SF\ satisfies dialectical monotonicity iff 
                \!\!\!${\Atts}(\arga) \!<\! {\Atts}(\argb)\!\wedge\! {\Supps}(\arga) \!=\! {\Supps}(\argb)\rightarrow\SF(\arga) \!>\! \SF(\argb)$;\!\!\!          \\
         &
     \!\!\!${\Atts}(\arga) \!=\! {\Atts}(\argb)\!\wedge\! {\Supps}(\arga) \!<\! {\Supps}(\argb)\rightarrow\SF(\arga) \!<\! \SF(\argb)$;\!\!\!
     \\
     &
     \!\!\!${\Atts}(\arga) \!=\! {\Atts}(\argb)\!\wedge\! {\Supps}(\arga) \!=\! {\Supps}(\argb)\rightarrow\SF(\arga) \!=\! \SF(\argb)$.\!\!\!
     \\
     \hline
     \!\!\!\emph{Additive}\!\!\! & $\forall \arga\in \Args$: \\
     \!\!\!\emph{Monotonicity}\!\!\! &  $\SF(\arga) = \sum_{%(\argb, \arga) \in \Supps
     \argb\in \Supps(\arga)}{\SF(\argb)} - \sum_{
		%(\argb, \arga) \in \Atts
		\argb \in \Atts(\arga)}{\SF(\argb)}$.
		\\
     \hline
\end{tabular}
    \caption{Some dialectical properties for %strength 
    $\SF$ when using BAFs $\langle \Args, \Atts, \Supps \rangle$. We adopt notations as follows.  (i) $\forall \arga\! \in \!\Args$ and $\Rels_i\!\in\! \{\Atts, \!\Supps\}$: ${\Rels_i}(\arga) \!=\! \{ \argb | (\argb, \arga) \!\in\! \Rels_i \}$.      (ii) $\forall A_1, A_2 \!\subseteq \!\Args$: $A_1 \!\leq\! A_2$ iff $\exists$ injective mapping $m:\!A_1 \!\mapsto \!A_2$ such that   $\forall \arga \!\in \!A_1$,  $\SF(\arga) \!\leq \!\SF(m(\arga))$;  $A_1 \!<\! A_2$ iff $A_1 \!\leq \!A_2\!\wedge \!A_2 \!\not\leq \!A_1$; $A_1 \!= \!A_2$ iff $A_1 \!\leq \!A_2\!\wedge \!A_2 \!\leq \!A_1$.  }
    \label{tab:props}
\end{table}
%%%%%%%%%
A natural candidate for inclusion in $\Pi$ %in the running example 
is \emph{dialectical monotonicity} (see formalisation in Table~\ref{tab:props} for BAFs, adapted from \cite{Baroni_18}), intuitively requiring that attacks weaken arguments and supports strengthen arguments. 
%, requiring that the strengthening of any argument will result in the weakening (strengthening) of any argument it attacks (supports, \resp).
Another is  
\emph{additive monotonicity} (see formalisation in Table~\ref{tab:props}, again for BAFs), requiring that \AR{an} argument's strength amounts to the sum of the strengths of \del{their }\AR{its }supporting arguments and the negations of the strengths of \del{their }\AR{its }attacking arguments.
% \footnote{     TOO MUCH OF A DETAIL - ALSO REALLY WE SHOULD HAVE THE BIASES AS ARGUMENTS ALREADY....\del{Notice that in this formulation bias arguments and the corresponding neurons were not included for simplicity, though necessary for additive monotonicity to hold.}}}

\begin{definition}
Let $G\!=\!\langle \Args, \Rels_1, \ldots, \Rels_m \rangle$ be the extracted GAF given choices \nos{(for hyper-parameters)}
 $\rep$, $t_1, \ldots, t_m$ and $\rc_1, \ldots, \rc_m$. Let \nos{hyper-parameters} $\SF$ and $\Pi$ be given.  
Then, $G$  is  \emph{$\Pi$-compliant under $\SF$} iff $\SF$  satisfies $\Pi$ in $G$. 
\end{definition}
%
%\begin{example}[Hyper-parameter $\Pi$]
%\label{ex:Pi}

\begin{figure*}[!ht]
    \centering
	{\includegraphics[width=1\textwidth]{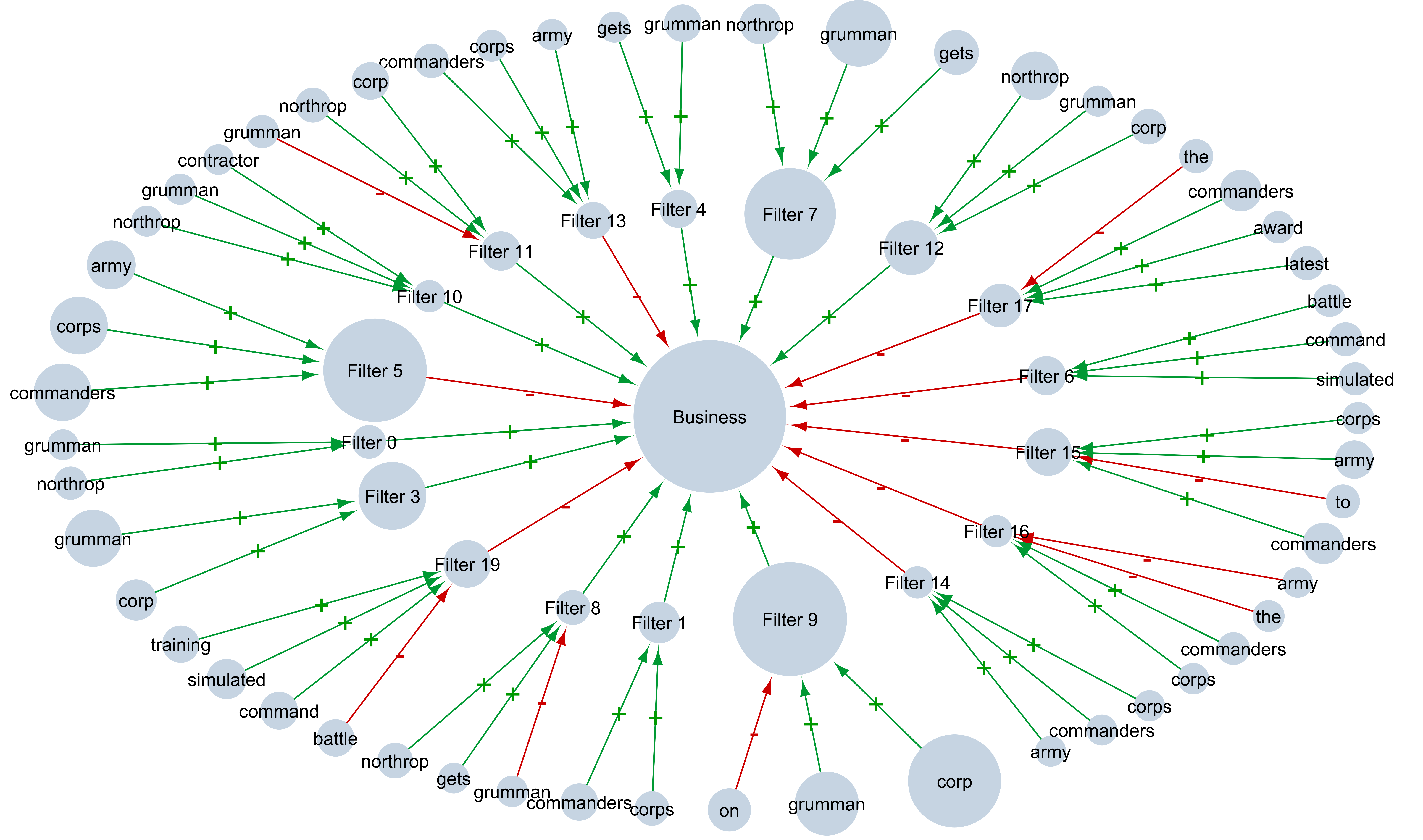}}
	\caption{%\delx{Bipolar argumentation framework} 
	\FTx{BAF} from which the DAX for text CNN in Figure~\ref{fig:dax_text} is %obtained
	drawn. %The bubbles  nodes are arguments 
	For each argument $\arga% \in \Args
	$, %with sizes corresponding to their %dialectical 
	size corresponds to 
	strength $\SF(\arga)$ and %as label their mapping to nodes
	label corresponds to mapping to node $\rho(\arga)$. Green (+) and red (-) arrows  indicate, \resp, support and attack.}
    \label{fig:af_text}
\end{figure*}

In the running example, 
 with $\SF(\alpha)$ given by the modulus of $\rep(\alpha)$'s activation %(see Example~\ref{ex-sigma}),
dialectical monotonicity equates to the requirement that increasing the magnitude of a neuron's  activation  results in a decrease (an increase) in magnitude of the activation of any neuron it attacks (supports, \resp) -- and it is thus desirable for explanation of outputs in terms of attacks and supports. 
The BAF in the running example is \del{$\Pi$-compliant for $\Pi$ amounting to }dialectical monotonicity\AR{-compliant} %In general, dialectical monotoniticy can hold 
with \FTkdd{some} activation functions%; in the running example it  can be seen by inspection that 
, e.g., it holds if \emph{tanh} %\delete{activation functions are}
is used \AR{since for a given connection with weight $w$ from activation $a_1$ towards a \emph{tanh} activation $a_2$, increasing (decreasing) $w a_1$ increases (decreases, \resp) $a_2$.}%\EAB{, it holds}
.
%\todo{AR what about additive monotonicity? does it hold? NO BECAUSE THERE ARE WEIGHTS AND BIASES is it desirable? NO BECAUSE THE NN WOULD BE TOO PRIMITIVE I THINK?}
\FTkdd{\AR{However}, the BAF is not additive monotonicity-compliant, given that $\SF$ disregards \AR{connection} weights in $\NN$.}
%\end{example}

%The satisfaction of such dialectical properties becomes more complicated as more realistic networks are used, e.g. since strata may not be adjacent and activations calculations will be more complicated, and so measures of strength other that the activations may be required. It can thus be seen how the choice of dialectical strength and relation characterisations is critical, and they can constructed so that the extracted argumentation framework may satisfy dialectical properties.

Several concrete choices (for hyper-parameters) 
may be possible and natural at Step 2, depending on the setting of deployment, the underlying \nos{choice of} NN and the choice of (hyper-parameter) $\sneurons$ at Step 1, as we will see in Section~\ref{sec:instances}. 
\nos{These choices need to be performed simultaneously as they depend on one another.}
Note %also 
that the choices of $\Pi$ and $\SF$ are crucial to obtain dialectically meaningful \DAX s at Step 3 (from extracted dialectical property-compliant GAFs under the chosen strength)\nos{, and distinguish \AR{our approach}, in particular, from works on visualisation of NNs}.

\emph{Step 3: from (dialectically compliant) GAFs to \DAX s.}
This amounts to 
%Once we have extracted an argumentation framework representing the dialectical reasoning for an output, this may be used to generate 
extracting, from the GAF and notion of strength from Step 2, a suitable \DAX\ for the intended users. 
GAFs per se will typically not be suitable, \del{in the first place }\AR{firstly }because their size may be (cognitively) prohibitive to users (as we shall see in Section~\ref{sec:instances}),
but also because, even when of manageable size, 
their outlook may not suit users' %explanatory 
requirements for the settings of deployment of explanations.
% Even when the size of GAFs is manageable (as in Figure~\ref{fig:eg}) users may need specific \emph{formats}  for explanations  to suit their explanatory requirements for the settings of deployment of explanations. 
%%%
%While the information content for \DAX s is harboured in the GAF, different \emph{formats} $\phi$ may suit different explanatory requirements. 
%%%
The choice  of \DAX 's \emph{format}, amounting to how much of the GAF from Step 2 to show users \del{and }\AR{in which way}, is determined by hyper-parameter  $\phi$, which, 
while guaranteeing that \DAX s retain an argumentative outlook,
takes into account the users' needs. 
In our running example, Figure~\ref{fig:eg}iii,
 $\phi$ selects only the ``strongest'' (according to $\SF$) attackers and supporters in the GAF and delivers a  
\emph{conversational} \DAX. 

Note that the selection of just some arguments (and dialectical relations involving them) matches other explanation methods, e.g.,  LIME, where only some features may be included to ensure that explanations are comprehensible to users.
%We envisage that the size of $\Args$ will depend on the needs of users, e.g., expert developers may benefit from larger $\Args$ than lay users. 
Also, conversational explanations (as in Figure~\ref{fig:eg}iii) are advocated by several, e.g., in \cite{Balog_19%,RT
}. For our instances in Section~\ref{sec:instances} we will explore, as part of $\phi$,  (interactive)
\emph{graphical} \DAX s (e.g., as in Figure~\ref{fig:dax_text}% for a CNN-based text classifier
, discussed later). 

Finally, 
users at the receiving end of explanations may also
require appropriate \emph{interpretations} of arguments and dialectical relations,
% e.g.\FTa{,} as word clouds in Figure~\ref{fig:dax_text}, 
returned by a suitably defined mapping $\user$ (left unspecified in Figure~\ref{fig:eg}iii, see Section~\ref{sec:instances} for some examples in our chosen instances).
The definition of $\chi$ may benefit from work on explanation-as-visualisation (e.g., as in \cite{Olah2017,Bau_17} for images). However, for \DAX s, this interpretation only plays a role in the last step, on top of the GAF obtained from the first two.

\section{Deploying the \DAX\ framework}
\label{sec:instances}

We %applied the general DAX methodology 
show how to deploy the \DAX\ framework with 
three specific neural architectures for prediction, trained on %specific 
datasets/tasks as indicated.   We give %\delx{full} 
details for the first, and sketch the rest (further details %{for the three instances} 
can be found in Appendix A% alongside illustrations of GAFs for these settings, including as many as 1,472,224 arguments
), focusing instead on motivating the underpinning choices for the hyper-parameters. 

\textbf{\DAX s for CNNs for Text Classification.}
\label{CNN-text}
We targeted the CNN architecture  in
\cite{Kim_14_TextCNN}
with an input layer of 150 words followed by: an embeddings layer (6B GloVe %\cite{GLOVE}
embeddings of size 300); a hidden layer of 20 1D ReLU convolutional filters of which 10 %are 
of filter size 3, 5 %are 
of size 2 and 5 %are 
of size 4, followed by a max pooling layer; and finally a dense %\del{layer with a softmax activation} 
softmax layer.
We trained this architecture 
with two datasets: a pre-processed version of \emph{AG-News} \cite{Gulli} (without HTML characters and punctuation) and \emph{IMDB} \cite{imdb}, in both cases 
 tokenizing sentences using the \emph{spaCy} tokenizer. In the remainder of %\delx{this part} 
%Section~\ref{CNN-text} 
\FTkdd{this section} we assume as given \NN\  for either dataset.

\emph{Step 1}.
% we choose to decose the CNN $f$ into two \textit{influence functions}: $\f_1$ from the input (stratum 0, $l_0 = 150$) to the intermediate features outputted by the max pooling layer (stratum 1, $l_1 = 20$) and $\f_2$ from the intermediate features to the prediction vector (stratum 2 with $l_2 %= n_p
% $ %, where $n_p$ is 
% equal to the number of predicted classes depending on the classification problem and corresponding dataset).
% Therefore, in this neural network the influences induced by the \textit{influence functions} link the input words at stratum 0 to the intermediate features at stratum 1, and these intermediate features to the prediction probabilities at stratum 2.
We %choose to have 
chose hyper-parameter $\sneurons$ with 
3 strata: %an input stratum 
$\sneurons_1$ with nodes corresponding to the input words ($|\sneurons_1| = 150$), %an intermediate stratum 
$\sneurons_2$ with nodes corresponding to the neurons of the max-pooling layer ($|\sneurons_2| = 20$) and %an output stratum 
$\sneurons_3 = \{ n_{\out} \}$ where $n_\out$ %is the output node corresponding to 
is
the neuron of the most probable class%($|\sneurons_3| = 1$)%\del{ before the softmax activation}
. 
Influences were then obtained as 
%defined in Section~\ref{sec:methodology}.
per Definition~\ref{def:influences}.
Thus, the resulting influence graph mirrors closely the structure of the underlying \NN.

%%%%%%%%%%%%%%%%%%
% In step 2, we have to characterise the property that the AEF satisfies in order to extract a BF.
% The property upon which the AEF extracts the BF is based on the LRP-0 \cite{Bach_15} (Layer-wise Relevance Propagation) rule, and specifically it considers the relevance back-propagated from one neuron to another. }

\begin{figure*}[ht]
    \centering
    \includegraphics[width=1\textwidth]{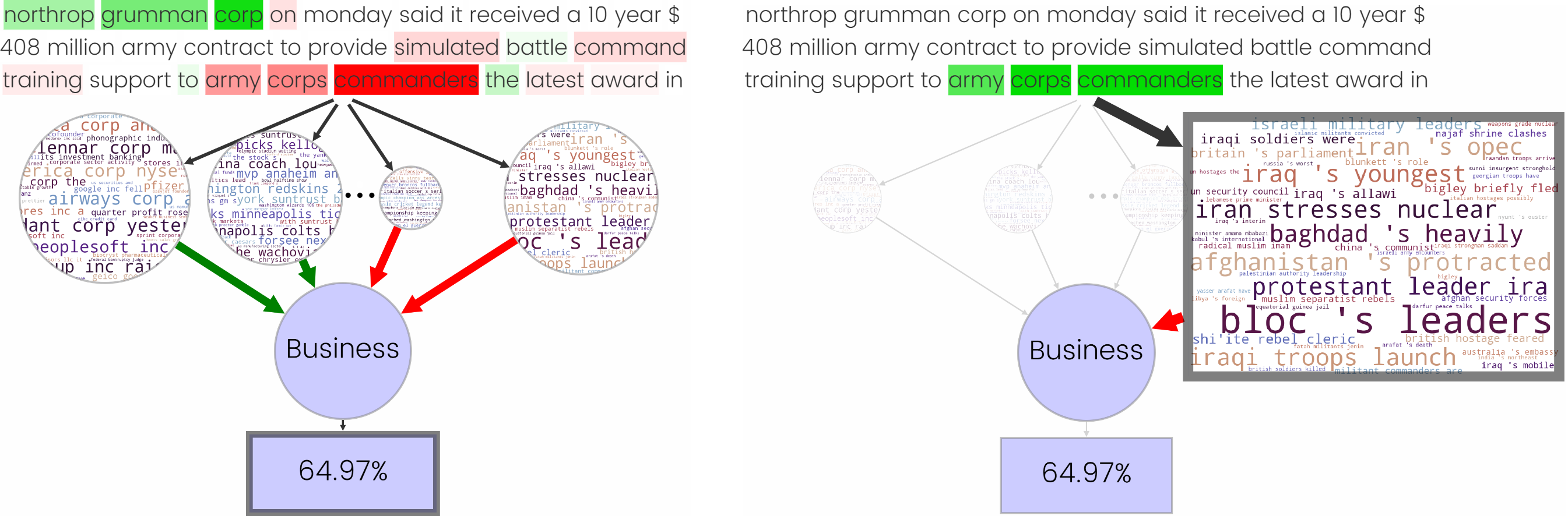}
    \caption{
    %\del{Graphical interactive DAX explanation for text classification with arguments on three levels: \textit{input words} with highlighting corresponding to the sum of the strengths of their arguments, the \textit{convolutional filters} with size representing their strength, the argument of the most probable \textit{class} and its probability. The colours green and red indicate, resp., support and attack.}
	    (Left) Graphical %\delx{interactive} 
	    DAX for %\delx{text classification}  
	    the (correct) output class ``Business'' (determined with 64.97\% probability) by the CNN for text classification with AG-News\nos{ from Section~\ref{CNN-text}}, for the input article from AG-News at the top. 
	    %\delx{in its initial state (left) and}
	(Right) The view of the DAX after clicking on the rightmost argument/word cloud%\delx{(right)}
	.  
	%For this DAX, arguments are on 3 levels, 
	Arguments are visualised (by $\chi$) as follows: \emph{input words} with highlighting of a word %corresponding to 
	given, on the left, by the sum of the (dialectical) strengths of all arguments representing that word and, on the right, by its corresponding argument's dialectical strength; \emph{convolutional filters/word clouds} with sizes %representing 
	the corresponding arguments' strengths; and the most probable \emph{class} and its probability. The  green/red colours indicate, \resp, support/attack (\nos{in both images, %colours are 
	}used on edges/relations from intermediate to output arguments, and on words from input to intermediate arguments).}
    \label{fig:dax_text}
\end{figure*}
\emph{Step 2}. 
We chose to extract a BAF $\langle \Args, \Atts\!, \Supps \rangle %\in \GFAll
$ with dialectical relations $\Atts %\!\subseteq\! \Args \!\times\! \Args
$ and
$\Supps %\!\subseteq\! \Args \!\times\! \Args
$ %represent 
of types attack (i.e., negative influence) and support (i.e., positive influence, \resp)\FTkdd{, to show evidence and counter-evidence, \resp, for predictions}.
\FTkdd{An example such BAF  (for a given input-output pair, with \NN\ trained on AG-News) is shown in Figure \ref{fig:af_text}. 
}
%\todo{EA/PL/AR why this choice???}
%\del{We will thus generate an intuitive DAX based on negative influences (attacks) and positive influences (supports) between nodes.}
$\Args$ is divided in 3 disjoint sets, matching the strata:
\ $\Args_3 = \{ \arga_\out \}$, with $\arga_\out$ the   \emph{output argument}; 
% $\Args_{2} \subseteq \{ \arga_{j} : j \in \{ 0, 1, ..., |\sneurons_2| \} \}$ is the set of \textit{intermediate arguments}, each of them corresponding to the features in the the second stratum;\\
% $\Args_{1} \subseteq \{ \arga_{ij} : j \in \{ 0, 1, ..., |\sneurons_2| \}, i \in \{ 0, 1, ..., |\sneurons_1| \}\}$ is the set of \textit{input arguments}, each one of which corresponds to a word in the first stratum and an intermediate feature in the second stratum.
$\Args_{2}$ is the set of all \emph{intermediate arguments} $\arga_{j}$ representing filters $n_{j}$ in %the second stratum
$\sneurons_2$; and
$\Args_{1}$ is the set of all \emph{input arguments} $\arga_{ij}$, each representing a word $n_{i}$ in %the first stratum 
$\sneurons_1$ that influences the  filter $n_{j}$ in %the second stratum
$\sneurons_2$.
The mapping from arguments to nodes is thus defined, trivially, as
$\rep(\arga)=n_i \in \sneurons_1$ if $\arga = \arga_{ij} \in \Args_1$, 
$\rep(\arga)=n_j \in \sneurons_2$ if $\arga = \arga_j \in \Args_2$ and 
$\rep(\arga)=n_\out$ if $\arga = \arga_\out \in \Args_3$.
%
%\EA{$\rep(\arga) = \begin{cases}
%n_i \in \sneurons_3 & \text{if } \arga = \arga_{ij} \in \Args_3\\
%n_j \in \sneurons_2 & \text{if } \arga = \arga_j \in \Args_2\\
%n_o & \text{if } \arga = \arga_o \in \Args_1
%\end{cases}$}
%
Even though in this DAX instantiation we chose to use 
the same types as in the toy illustration in Section~\ref{sec:methodology}, %in this setting 
here
we %choose to 
based %the definition of 
the relation characterisations $\rc_+$, $\rc_-$ %\delx{and 
%the dialectical strength $\SF$} 
on LRP-0~\cite{Bach_15}. 
Formally, for  $R(j,i)$ %stand for LRP \emph{relevance} 
the LRP-0 relevance back-propagated from $n_j$ back to $n_i$:\footnote{If $n_i$ and/or $n_j$ are groups of neurons then 
%the relevance back-propagated from \nos{node} $n_j$ to \nos{node} $n_i$ 
$R(j,i)$ is the sum of the relevances back-propagated from every neuron in $n_j$ to every neuron in $n_i$.} $c_{-}(n_i, n_j) \!=\! true$ iff $R(j,i) \!<\! 0$ and $c_{+}(n_i, n_j) \!=\! true$ iff $R(j,i) \!>\! 0$. 
%\del{ $\quad\quad\quad\quad$$c_{-}(n_i, n_j) = 
%% \begin{cases}
%true \quad \text{iff }  R(j,i) < 0
%% false & \text{otherwise}
%% \end{cases}
%$
%$\quad\quad\quad$
%$c_{+}(n_i, n_j) = 
%% \begin{cases}
%true \quad \text{iff }  R(j,i) > 0
%% false & \text{otherwise}
%% \end{cases}
%$}
%%%
We also defined the dialectical strength $\SF$ in terms of LRP-0: for
$a_{p}$ the activation of neuron $n_p$,
$\SF(\alpha)\!=\!a_\out$ if $\alpha\!=\!\arga_\out\!\in\!\Args_3$, 
$\SF(\alpha)\!=\!|R(\out,j)|$ if $\alpha\!=\!\arga_j\!\in\!\Args_2$ and
$\SF(\alpha)\!=\!\left|R(j,i) \cdot \frac{R(\out,j)}{a_j} \right|$ if $\alpha\!=\!\arga_{ij}\!\in\!\Args_1$.
%%%%%%%%%%%%%%
% $\quad\quad \quad \quad  \SF(\alpha) = \begin{cases}
%     a_\out& \text{if } \alpha = \arga_\out \in \Args_3 \\
%     |R(\out,j)| & \text{if } \alpha = \arga_j \in \Args_2 \\
%     \left|R(j,i) \cdot \frac{R(\out,j)}{a_j} \right| & \text{if } \alpha = \arga_{ij} \in \Args_1
% \end{cases}$
%
\FTkdd{We chose LRP because this is a robust, efficiently computable interpretability method. Also,}
%
%In the choice of  $c_{-}$, $c_{+}$, and $\SF$, 
we were driven by \FTkdd{compliance with} \nos{the properties of} 
\emph{dialectical monotonicity} and 
\emph{additive monotonicity}  from Table~\ref{tab:props}. 
These properties %\delete{characterise} 
lead to BAFs from which  \DAX s aligning with human judgement can be drawn, as we will show in Section \ref{sec:humans}. 
%
%A BAF extracted at this step (for a given input-output pair, with \NN\ trained on AG-News) is shown in Figure \ref{fig:af_text}. 

\emph{Step 3.}
The BAFs %from step 2 
may be quite large (e.g., the one in  Figure \ref{fig:af_text} includes  73 arguments). 
Also, their arguments (and the ones representing filters in particular), are unlikely to appeal to users (the intermediate %arguments/filters 
ones
are basically ``black-boxes'')\nos{; indeed, as mentioned in Section \ref{sec:methodology}, these BAFs are not meant to be shown to users}.
We chose to generate \DAX s with only \EABC{the top 3 supporters and attackers} and in \emph{graphical, interactive} format ($\phi$), as shown in Figure~\ref{fig:dax_text} (for 
%a specific input text and relative CNN-computed output, with \NN\ trained on AG-News
the same setting as in Figure~\ref{fig:af_text}\EABC{, but limited to 4 arguments for space constraints}).
The interactive format of this \DAX\ allows users to focus on particular parts \FTkdd{(by clicking)}: this is especially useful \nos{with deep explanations,} to help users control the amount of information they receive.%

The \DAX s in this instance have 3 levels, aligning with the BAFs.
Since convolutional filters/intermediate arguments are not intrinsically %\del{associated with} 
human-comprehensible, we opted (in the  choice of mapping $\user$) to pair them with word clouds showing n-grams from input texts in samples in the training set that activate the most the corresponding filter, \FTkdd{with the n-grams' size reflecting the number of activating samples,} as in \cite{FIND}. 
%\nos{In this way, users could relate these arguments with a human-understandable meaning in the dialectical debate that the AFs generate.} 
We also chose to visualise attacks and supports from the intermediate arguments with %differently coloured 
red and green (\resp.) arrows, and the attacks and supports from the input to the intermediate arguments (word clouds) \FTkdd{diffently in the static and interactive presentation of \DAX s. In particular,  when the word clouds are selected during interactions with the user, as in Figure~\ref{fig:dax_text}, right (resulting from a user clicking on a word cloud, leading to this being magnified and the single words supporting or attacking it being highlighted), attacks and supports are arrows originating from %differently coloured 
red and green (\resp.) words. Instead, when \DAX s are shown in their initial, static form (as in Figure~\ref{fig:dax_text}, left) the information about attacks and supports from input arguments (words) is reflected in the colours and intensity of the words, so that} \EABC{green and red represents words that overall support or attack the prediction, \resp, and intensity the magnitude of their contribution}.

\textbf{\DAX s for CNNs for Image Classification.}
% }
\label{CNN-image}
%The sphere of application of DAX is \AR{not} limited to neural network with only one hidden layer, but it is suited also \AR{to} deep neural networks. 
To assess \DAX's applicability to deeper NNs, we considered a version of \textit{VGG16} \cite{Simonyan2014} pre-trained on \textit{ImageNet} \cite{imagenet_cvpr09}, %classifying an image amongst 
with 1000
%\del{alternative }
classes%. This NN comprises
, 5 blocks of convolutional layers, % \del{, each of them followed by} 
a max pooling \nos{layer,} 
and 3 fully connected layers.
%As in Section \ref{CNN-text}, we will now walk trough each of the steps outlined in Section \ref{sec:methodology} to deploy DAX in this setting.
%%%%%%%%%%%

In \emph{Step 1}, %This \NN\ has many layers. 
to avoid %\delete{the generation of} 
%explanations that may overwhelm 
overwhelming users with information, we chose to `promote' only 3 of the NN's layers to strata: an input stratum $\sneurons_{1}$ whose nodes correspond to pixels in input images (combining 3 neurons each, for the pixels' 3 RGB channels),   %($|\sneurons_1| = 224 \cdot 224$),
an intermediate stratum $\sneurons_2$ whose nodes correspond to filters in the last convolutional layer (with each node a collection of 14x14 neurons) %($|\sneurons_2| = 512$),
and an output stratum $\sneurons_{3} \!= \!\{ n_\out \}$ 
%($|\sneurons_3| \!=\! 1$) where $n_\out$ %is the output node, corresponding to 
%is 
with $n_\out$ the (single) neuron of the most probable (output) class. Overall, this gives $|\sneurons| \!=\! 224 \cdot 224 + 512+ 1$. \FTkdd{We chose filters as members of $\sneurons_2$  because they are known to naturally correspond to ``concepts''  amenable to intuitive visualisation based on activation maximization  \cite{Olah2017,Kotikalapudi2017%\nos{,Mahendran2015,Mordvintsev2015
} by $\chi$ in \emph{Step 3}.}

\begin{figure*}[!ht]
    \centering
    \includegraphics[width=1.0\textwidth]{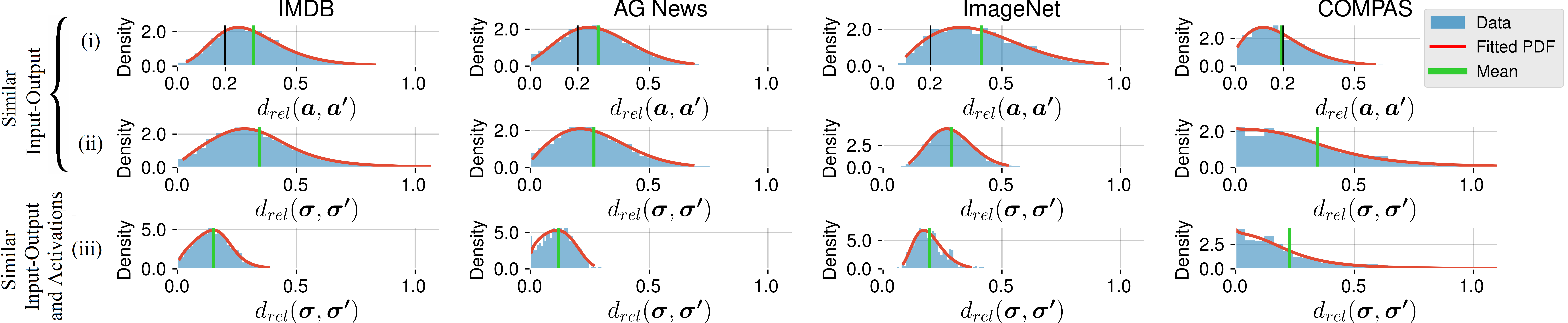}
    \caption{
    % Distributions of difference of the intermediate arguments' \EABC{activations and} strengths, i.e., \EABC{$d(\bm{x}, \bm{x}') = \frac{2 \cdot ||\bm{x'} - \bm{x}||_1}{||\bm{x'}||_1 + ||\bm{x}||_1}$ where $\bm{x}$ and  $\bm{x'}$} are \resp\ the strengths' vectors of the arguments extracted from the original input and the similar one. 
    \EABC{Distributions of difference of %the 
    (i) intermediate strata activations and (ii)  intermediate arguments' strengths, %extracted from the original input and that with similar input and output
    \FTkdd{for given and \emph{similar} input-output pairs}, and (iii) %difference in 
    \FTkdd{of} strengths of \emph{similar} input-output pairs that also have \emph{similar} activations.
    %The differences are relative, i.e., 
    \FTkdd{Here, for $\bm{x}$ and  $\bm{x'}$} two vectors (of activations or strengths)% extracted from the original sample and the similar one
    , the relative difference between $\bm{x}$ and  $\bm{x'}$ is} \EABC{$d_{rel}(\bm{x}, \bm{x}') = \frac{2 \cdot ||\bm{x'} - \bm{x}||_1}{||\bm{x'}||_1 + ||\bm{x}||_1}$%where $\bm{x}$ and  $\bm{x'}$} are two vectors (activations or strengths) extracted from the original sample and the similar one
    .}
    % We show results for the (NNs for) IMDB (i), AG-News (ii)%\nos{ datasets on CNN for text classification}
    % , a sample of ImageNet (iii), and  COMPAS (iv)
    %\todo{If we like it: change median to mean}.
    }
    \label{fig:empirical}
\end{figure*}

In \emph{Step 2} we chose a GAF with \nos{arguments matching the nodes and} a single relation of support (thus the GAF is a SAF, see Section~\ref{sec:background}), %and contrary to the instantation in Section \ref{CNN-text}, 
%since in this context we are dealing with a CNN for image classification, we chose 
and used Grad-CAM weighted forward activation maps \cite{Selvaraju2016} %of the last convolutional layer 
to define relation characterisations and dialectical strength. 
%\nos{The set of arguments $\Args=\Args_1 \cup \Args_2\cup\Args_3$, representing the nodes in the strata ($\Args_3 = \{ \arga_\out \}$, with the \emph{output argument}, $\Args_{2}$ with \emph{intermediate arguments} representing filters in the the second stratum, and \emph{input argument} $\arga_{xyj}$ in $\Args_{1}$ representing  pixel %\todo{WHAT IS $xy$? Pixel coordinates in the image} in the first stratum and the filter $j$ in the second stratum.}
%
%Note that, we chose to 
%The focus on the support dialectical relation only and the exclusion of attack) is in line with Grad-CAM, whose focus is on features \emph{positively} influencing predicted classes.
\FTkdd{We chose to exclude attack for simplicity, and because \AR{it does} not naturally play\del{ing} a role in ImageNet. The choice of Grad-CAM, rather than LRP, is in line with the focus on support, given that it identifies features \emph{positively} influencing predicted classes.
To define relation characterisation and strength, we applied Grad-CAM not only to the ``target concepts'', i.e., the classes, but also to the intermediate convolutional nodes/arguments. These choices give compliance with    \emph{dialectical monotonicity} and \emph{additive monotonicity} properties (amounting, for SAFs, to removing all terms involving attack in the formulations in Table~\ref{tab:props}).}
%%%%%%%%%%%%%%%%%%%%%

\FTkdd{Finally, at \emph{Step 3}, we may choose a variety of formats, including a similar \nos{graphical, interactive }format as for the text-CNN (see Appendix A), with (a suitable selection of) %the 
intermediate arguments visualised (by $\chi$) using activation maximisation, as discussed earlier. }

%\textbf{Step 3.} We obtained (from the SAF) a \textit{graphical}  \DAX\ on 3 levels: the input image, the intermediate strata's filters and  the predicted class node%\nos{ \EA{with} its probability}
%, in the spirit of the text-CNN. %in Section \ref{CNN-text}, 
%Here, though, to give a human-comprehensible meaning to the  intermediate arguments, we  defined $\user$  in terms of feature visualisation based on activation maximisation \cite{Olah2017,Kotikalapudi2017,Mahendran2015,Mordvintsev2015}. %\todo{A picture would be lovely ...}

\textbf{\DAX s for %\feedforward\ neural network 
FFNNs %for binary classification 
with tabular data.
}
\label{FF}
\FTkdd{As third instance, we focused on NNs for tabular data, for which the identification of suitable explanation can be challenging
\cite{kdd2020}.
%focuses on explanations for predictions by NNs indicating how difficult it is to show explanations to users for this type of data....so we had to be creative...also why tripolar? counterfactual
Specifically,}
%In this section we will explore how the DAX general methodology can be applied to extract a broader spectrum of relations, and in particular we will focus on the extraction relations characterized as counter-factual.
%
%To explore a broader spectrum of dialectical relations (including a third relation of \emph{strict support} of a counterfactual nature), 
we deployed an FFNN  for classification with the (categorical) %tabular data in the 
COMPAS dataset \cite{ProRepublica2016}, understanding
\iffalse
\nos{\textit{sex},
\textit{age},
\textit{race},
\textit{juv\_fel\_count},
\textit{juv\_misd\_count},
\textit{juv\_other\_count},
\textit{priors\_count},
\textit{is\_recid},
\textit{is\_violent\_recid},
\textit{custody},
\textit{charge\_desc},
\textit{charge\_degree}
as categorical input features 
and}
\fi
\textit{two\_year\_recid} as binary prediction %\footnote{Refer to the Supplementary Material for details\delete{, including on pre-processing}.}
(see Appendix A
%\FTx{Appendix~\ref{app:FF}} 
for details).
% The trained network has a prediction accuracy of 70\%, that even though low it's not the focus of this paper.
The FFNN 
has
3 layers: an input layer that takes the one-hot encoding of the inputs, a dense hidden layer %\del{with 8 neurons}
of size 8 with $tanh$ activation followed by %\del{a }
$ReLU$ and an output layer with $sigmoid$ activation.
%%%%%%%%%%%%%%%
%We will now walk trough each of of the steps outlined in Section \ref{sec:methodology} to deploy DAX in this context.

In 
\emph{Step 1}, \FTkdd{we chose to keep as much as possible of the (small) NN; thus} the first two layers coincide with the respective strata and the last layer amounts to 
the most probable class node $n_\out$ from the last stratum,
giving 
$|\sneurons| =58+8+1$.
\iffalse
%stratum 
$\sneurons_1$ ($|\sneurons_1| = 58$){ with $n_f \in \sneurons_1$ corresponding to  input neuron $f$}% 
% \footnote{Here each node is the group of neurons of feature $f$}
, %a middle stratum 
$\sneurons_2$ such that $n_j \in \sneurons_2$ is the node corresponding to the hidden neuron $j$ after (tanh and) ReLU activation ($|\sneurons_2| = 8$), and %an output stratum 
$\sneurons_3 = \{ n_\out \}$ with $n_\out$ %is the output neuron of 
the most probable class node.
\fi

In \emph{Step 2},
\FTkdd{we opted for a form of GAF able to unearth evidence and counter-evidence (in the spirit of %Section~\ref{CNN-text}
\FTkdd{the text-CNN}), while being able to represent \emph{counter-factuality}, useful to identify possible biases in \NN\ (often a concern with classifiers for COMPAS% \todo{for AR: any reference we can add? our BC? better  somebody else for whom we already have a reference we could use without adding more?}
). 
%in the spirit of \cite{BC}). 
} 
The GAF is a TAF $\langle \Args, \Atts, \pSupps, \Rels^{+!} \rangle$ with \nos{each node in \sneurons\ represented by at least one argument in \Args\ (as per Definition~\ref{def:args}) and }dialectical relations, \resp, of attack, support and \emph{critical support}, \FTkdd{drawn from} essential positive influences \FTkdd{ removing which would lead to different predictions}. %
\FTkdd{We chose to define strength and relation characterisations in terms of activations (as for the toy illustration in Section~\ref{sec:methodology}), as this reflects well our desired counter-factual behaviour of $\Rels^{+!}$:}
%Here, the drivers of the extraction of the TAF (dialectical strength and relations) are the activations.
intuitively, \nos{the relation characterisation}
$\rc_{+!}$ defines a critical support between arguments representing $n_i$ and $n_j$ iff deactivating $n_i$ causes the deactivation of $n_j$ (i.e., %neuron 
%$n_i$ is not only positively influencing $n_j$, but its 
$n_i$'s positive contribution is essential for %\delete{the activation of neuron $n_j$} 
%its 
$n_j$'s activation).
%The dialectical property chosen here is of \emph{binary counter-factuality}, 
\FTkdd{These choices lead to compliance with the dialectical property of}
\emph{%binary 
counter-factuality}, requiring that the removal of critical support from an argument to another with positive dialectical strength will result in the strength of the latter becoming zero or lower.

In \emph{Step 3} we %transform the TAF into a graphical explanation 
may again choose, amongst others, 
%chose again an  graphical DAX (obtained from the TAF)
a graphical, interactive \DAX\ with 3 levels reflecting the strata/argument structure\nos{: (1) the input features (2) the hidden features (3) the prediction and its probability.}. \FTkdd{No matter the format, we need to take the (non-trivial) decision on how to}   
visualise (via $\user$)
the intermediate arguments/hidden neurons and the dialectical relations \FTkdd{so %that 
they are human-comprehensible}. \FTkdd{For example, we could choose \nos{to use }pie charts 
showing %, simultaneously, 
%at the same time, 
intermediate arguments and \AR{their incoming} attack\AR{s}, support\AR{s} and critical support\AR{s} %to them 
from input arguments\del{/features} (see Appendix A)}. % the input features activating the most for each of the hidden neurons/arguments. %\todo{I wish we could have a small picture!!}

\section{Empirical Evaluation}
\label{sec:empirical}

%We focus our empirical evaluation on 
We consider, for
the  \DAX\ instances in Section~\ref{sec:instances}, %along 
two dimensions: \emph{computational cost} (standard in the literature \cite{Sokol_20}) and a novel %property of 
\emph{\depth},  tailored  to deep explanations such as \DAX s. 
%standard properties  for evaluation of explanations (\emph{stability} and \emph{computational cost}) that are standard in the literature (e.g., see the overview in \cite{Sokol_20}) as well as a third dimension (\depth) which is tailored  to deep explanations such as DAXs.  

\iffalse
\textbf{Stability%\nos{ and Fidelity}
}
All instantiations  use deterministic methods (LRP, GradCAM, activations) which, given the same input/output pairs, return the same information. So, all GAFs are unique for each input-output pair, and DAXs are \emph{stable} \cite{Sokol_20}. 
% \emph{fidelity} {performance} at {the} words level is equivalent to LRP, i.e. masking the most relevant word is equivalent to masking the word (associated with arguments) with the highest aggregated ({by addition}) strength.}
\nos{%Further,  DAX' \emph{fidelity}  at the input level is equivalent to the fidelity performance of underlying methods, e.g. masking the most relevant word is equivalent to masking the word (associated with 
%???? (by addition) strength.  In other words, 
The \emph{fidelity} of DAX is directly connected with the fidelity of the underlying methods used to calculate strengths.}
\fi

\textbf{Computational cost}. 
Firstly note that, although the choice of $\langle \sneurons, \inff\rangle$ is tailored to explaining $\f(\x)\!=\!\out$, for specific input-output pairs $(\x,\! \out)$, the choice of candidate nodes in \sneurons\ from the hidden layers can be made a-priori, independently of any  pair. Also, the skeleton underpinning the construction of \DAX s relies exclusively on the given %neural architecture
NN, prior to training.
Thus, there is a single one-off cost for constructing \DAX s that can be shared across %\delete{instances and} 
pairs.
Secondly, the time complexity to calculate \DAX s is relatively small: there is a one time cost for %the mapping 
$\user$% (from arguments)
, but %after that
the cost to generate a single 
\DAX\ is comparable to the cost of a single prediction and single (for the FFNN instance) or multiple (for LRP and Grad-CAM in the CNN instances) back-propagation steps. The time cost depends \textit{linearly} on the %number of selected nodes 
size of \sneurons% at step 1
. %Concretely, in
\FTkdd{Formally, for the text-CNNs, given the time cost for a prediction $c_{f}$ ($1.7 \pm 0.4$ ms, in our experiments) and the cost to back-propagate the output to the inputs $c_{b}$ ($50.2 \pm 11.0$ ms, in our experiments) while we let $c$ be any other (constant or negligible) computational cost, the (one-time) generation of the word clouds %(a one-time cost for $\user$) 
using $n_S$ training samples (we used 1000 %to generate  
for %the example in 
Figure \ref{fig:dax_text}) has time complexity: 
%\\
%\hspace*{1cm}
$O(n_S \cdot (c_{f} + c_{b}) \cdot |\sneurons_2| + c)$. %\\
The time complexity  to generate a single \DAX\ is instead: 
%\\
%\hspace*{1cm}
$O(c_{f} + c_{b} \cdot |\sneurons_2| + c)$. %\\
}
%%%%%
\FTkdd{Details for the other instances are in Appendix A. Here, note that 
 the setup took up to %$2.35$ 
 \FTkdd{2:35} hours for the VGG16 instance %(for activation maximization) 
 while the time %taken 
to generate a single \DAX\ ranged from %less than
under $1 ms$ (for the FFNN) %up 
to $4.35 s$ (for VGG16)}.
%\nos{Thus, DAX has computational advantages over other explanation methods that require the generation and prediction of multiple (possibly thousands) samples to provide an explanation.}
%\nos{Formally, for the CNN for text, given the time cost for a prediction $c_{f}$ ($1.7 \pm 0.4$ ms, in our experiments) and that to back-propagate the output to the inputs $c_{b}$ ($50.2 \pm 11.0$ ms, in our experiments), the generation of the word clouds (a one-time cost for $\rep$) using $n_S$ training samples (we used 1000 to generate the example in Figure \ref{fig:dax_text}) has time complexity $O((c_{f} + c_{b}) \cdot |\sneurons_2| \cdot n_S log(n_S))$. The time complexity  to generate a single DAX is instead: $O(c_{f} + c_{b} \cdot |\sneurons_2| + c_{n} \cdot |\sneurons_1| \cdot |\sneurons_2|)$ where $c_{n}$ ($0.21 \pm 0.07$ ms, in our experiments) is the cost to generate a single node of the BAF.}

\FTkdd{\textbf{\Depth}. 
Fidelity of generated explanations to the model being explained is widely considered an important property for XAI methods (e.g., see \cite{Sokol_20}), but existing notions of fidelity are mostly tailored to the dominant view of ``flat'' explanations, for outputs given the inputs, and their underpinning assumptions are unsuitable for deep explanations. 
As an example, \cite{faithfulness} poses that explanations for similar inputs for which the underlying model computes similar outputs should in turn be similar. This  disregards the possibility that the model computes these outputs in completely different manners. 
Indeed, we assessed this empirically for our three settings, showing that id does not hold. We defined similarity as follows.
For text classification, we generated similar input pairs back-translating from French \cite{ribeiro-etal-2018-semantically} using Google Translate; for images we paired each image with a noisy counterpart (Gaussian noise of $std = 10$); for tabular data we paired each sample with another sample with one categorical feature changed. In all settings, we considered outputs to be similar when the predicted probability changed less than $5\%$.
\EABC{Our experiments show that activations of the intermediate strata could differ considerably when inputs and outputs are similar, with an average relative difference ranging between $19.2\%$ and $41.3\%$, as shown in Fig.~\ref{fig:empirical}i.}
\EABC{Our \DAX s are able to match this inner diversity, as shown in Fig.~\ref{fig:empirical}ii, where we note how also the intermediate arguments' strengths follow this behaviour of the model.}
Our novel notion of \emph{\depth}, suitable for evaluating \DAX s (and deep explanations in general), sanctions that 
    \emph{``deep explanations for similar inputs for which the underlying model \emph{similarly} computes similar outputs should in turn be similar''.}
\EABC{In order assess this property, we added an additional constraint for two samples to be regarded as similar, requiring that also the intermediate strata activations must be similar. Experimentally, we considered activations to be similar if their average relative distance was less than $20\%$ (see Fig.~\ref{fig:empirical} caption for the formal definition). As shown in Fig.\ref{fig:empirical}iii}, our explanations consistently reflect %the addition of 
this constraint, showing a reduction %in terms 
of the average %relative
difference between the strengths of the intermediate strata arguments
%between $32.2\%$ and $55.8\%$
of $32.2\%$-$55.8\%$.}

\section{Experimental Results}
\label{sec:humans}
%In this section we evaluate the instantiation of DAX from Section \ref{CNN-text} empirically and experimentally.
%In this section, we 
We conducted experiments with 72 %human 
participants on Amazon Mechanical Turk%\nos{ (MTurk) to demonstrate several desirable properties of DAX, specifically}
,
to assess: (A) whether \DAX s are comprehensible% to humans
; (B) whether \DAX s align with human judgement and (C) how desirable \DAX s are when compared to other explanation methods with an argumentative %flavour
spirit.
We 
focused on %\nos{the DAX instantiation for the CNN or} 
text classification \nos{(see Section \ref{CNN-text})}
with %the AG-News dataset
AG-News.
We chose this setting because topic classification (AG-News) is multiclass (4 classes) -- thus more complex than (binary) sentiment analysis (IMDB); also, it requires no domain expertise (unlike COMPAS) and has not been widely studied (unlike ImageNet). The participants were non-expert {(see Appendix C%\FTx{Appendix~\ref{app:human}}
}), showing that \DAX'\AR{s} target audience may %not be limited to experts
\FTkdd{be broad}. 

% We tested DAX on 72 MTurk workers with the aim to assess:
% \begin{enumerate}[label=(\Alph*),itemsep=0em]
%     \item whether DAX are \AR{comprehensible} to humans;
%     \item whether DAX \del{normally} \del{agree} \AR{align} with human judgement;
%     % \item that DAX intermediate strata provides information about the underlying classifier;
%     \item how desirable DAX are when compared to other explanation methods.
% \end{enumerate}

In the experiments, we %\delete{considered a} 
used only samples in the test set predicted with high confidence %\delete{if the} 
(i.e., %whose prediction had a 
with probability %greater than 
over $0.95$); we did so as, for samples predicted with lower confidence, if the explanations are implausible to users, we cannot know whether this %\delx{was}
is due to the poor underlying model or the explanations themselves.
\FTkdd{To obtain interpretations (by $\chi$) of intermediate arguments in \DAX s we used word clouds (see Section~\ref{CNN-text}) generated from 1000 random training samples. 
%Given the activations of the convolutional filters of 1000 random training samples, we 
We 
considered a \emph{filter} to be \emph{strongly activated} (by the input-output pair being explained) if its activation was greater than the 90th percentile of the activations of the 1000 training samples,} and \emph{weakly activated} if its activation was lower than the 1st percentile.
We considered a \emph{class} to be \emph{strongly supported} by an intermediate argument if 
the latter's strength ranked in the top 20\% of the arguments %associated with 
\FTkdd{supporting} the %former 
class.

% \del{the argument had a strength in the top 20\% of the range going from the minimum to the maximum strength of the arguments associated with the same filter.} %\PL{Can the last sentence be simplified?}

% \PL{(I think, to make the paper more readable, it might be a good idea to split the following contents into subsubsections based on the three (or four) items above: Understandability, Agreement with human judgements (+ intermediate strata), User Acceptance. But I found that the order of research questions 1-4 was rearranged. Agreements come before Understandability, so I'm not sure if subsubsections break any intended order.) }

% \todo{Not sure subsubsection vs. letters}
% \del{\subsubsection{Understandability}}
\textbf{(A) Comprehensibility.}
%\del{One novelty \AR{we propose is the use} of intermediate strata to disclose the inner decision making of the model.In particular, for  CNN text classifiers,} 
%\nos{We use word clouds to visualise intermediate arguments representing  convolutional filters (see Figure~\ref{fig:dax_text}).}
The following research questions
%\nos{, therefore, }
aim to ensure that humans can understand the interplay between intermediate arguments (i.e., word clouds) %, see Figure~\ref{fig:dax_text}) 
and other arguments  (i.e., input words, output classes).
%%%%%%%%%%%%%
% Task 2 (Former RQ2)
\textbf{RQ1}: \emph{``Can humans understand the \emph{roles} of  individual intermediate arguments towards the output %\nos{classes} 
by examining their word clouds?''} To answer RQ1, 
we showed participants a word cloud %interpreting an intermediate argument 
and asked them to select the class they %would 
best associated with it. 
\textbf{\textit{Results:}} Over $97\%$ \EABC{($\pm 0.3\%$)} of answers picked a class strongly supported by the presented argument, confirming human understanding.
%\nos{In }more than $97\%$ of the answers ($p << 0.001$)\nos{, participants} picked a class that is strongly supported by the filter, confirming that \del{they} humans understood the function of the intermediate \AR{arguments} in the classification.
%%%%%%%%%%%%%%%%%%%%
% Task 4 (Former RQ1)
\textbf{RQ2}: \emph{``Can humans understand the \emph{patterns} of intermediate arguments by examining their word clouds?''} To answer RQ2, we asked participants to select from 4 n-grams that which best matched the pattern in the word cloud of a strongly activated filter.
\del{For each question, we used an input text predicted with high confidence and a filter that is strongly activated:} The 4 n-grams were extracted from  input texts correctly predicted with high confidence by the CNN, and the CNN's best-match  was %that %\del{with the highest convolution score}
selected by the max pooling, while the others weakly activated the filter. 
\textbf{\textit{Results:}} \FTx{Participants answered correctly in $65.5\%$ \EABC{($\pm 1\%$)} of the cases}%
%\delx{$65.5\%$ correct answers}%\nos{, meaning that most of the participants understood the prominent characteristic of the n-grams which features were looking for}
. We posit this result as positive 
%since the answer is  
as %given that 
random guesses can lead to $25\%$ correct answers and the task is 
not trivial without NLP experience, as was the case for $96\%$ of participants %\FTx{(see Figure~\ref{fig:background} in Appendix~\ref{app:human})}
{(see Appendix C)}.
%%%%%%%%%%%%%%%
% \todo{Any better place for RQ5? Should it be a new subsubsection?}
% To verify that DAX intermediate strata provide information about the underlying classifier, we focused on the following research question.
%%%%%%%%%%%%%%%%%%%%
% Task 6 (Former RQ5)
\textbf{RQ3}: \emph{``Can humans understand the NN %\del{'s mechanism} 
from the dialectical relations originating from the intermediate arguments only, without the input%\nos{ text}
?''} %\del{ And how much information do we need to provide to the users?}''. 
To answer RQ3, %\del{We showed the DAX incrementally,} 
\nos{incrementally,} we first showed only the strongest supporter and %\nos{the strongest} 
attacker of the predicted class, and %subsequently 
then
added the next strongest  supporter and %\nos{next strongest} 
attacker, and so on until 4 of each were shown. At each step we asked the participants to select the class they would predict%\nos{ given this information}
.
% \del{As with other explanation methods, the amount of information in the output of DAX can be customized and tailored to a specific user base, we can in fact vary the amount of intermediate strata arguments showed to the user.}
\textbf{\textit{Results:}} %\del{Two word clouds} 
The strongest supporter and %\nos{the strongest} 
attacker %were sufficient 
sufficed for participants to correctly classify %the 
samples in %more than 
over $80\%$ of the cases. %\nos{, and 
% \del{increasing their number}
%adding more \del{word clouds} \AR{arguments} did not \AR{improve these results} \del{increase the correct predictions} significantly}. \del{This confirmed that the intermediate \AR{arguments} provide insightful information about the classifier, and}
Thus, even partial information about the intermediate arguments gives strong insight into the %\FTx{C}NN 
\FTx{model}.

% \del{\subsubsection{Agreement with Human Judgements}}
\textbf{(B) Alignment with Human Judgement.}
%\del{Alignment between machine and human judgement is \AR{critical for explanatory reasoning, and so we assessed} \del{important in several applications such as recommendations and debates. In this section, we want to check }}
We assessed 
whether \DAX s (especially %the notions of 
support% and attack
) %generally 
align with human judgement when the CNN confidently and correctly predicts a certain 
class.
%%%%%%%%%%%%%%%%%%%
% Task 1 (Former RQ4)
\textbf{RQ4}: \emph{``%
%\del{Given \del{that a classifier makes} a correct prediction with high confidence, }
Do %the notions of %attacks and 
supports \emph{between  intermediate %\del{strata} 
arguments %\nos{(associated with convolutional filters)} 
and predicted class} 
%in the generated explanation 
in \DAX s align with human judgement?''} 
To answer RQ4, 
we showed  n-grams extracted 
from  input texts correctly predicted with high confidence and selected by a strongly activated filter (but without showing the %corresponding 
word cloud) and asked  participants to select the class %with which 
this n-gram is best associated with. \textbf{\textit{Results:}} In %more than 
over $96\%$ \EABC{($\pm 0.4\%$)} of the answers %\del{ to RQ4}
, participants selected a class that 
the n-gram strongly supports. 
%\del{contributes positively towards.} \del{in DAX.}
 %%%%%%%%%%%%%%%%%%%%
% Task 3 (Former RQ3)
\textbf{RQ5}: \emph{``%\del{Given \del{that a classifier makes} a correct prediction with high confidence, }
Do %the notions of 
%attacks and 
supports \emph{between individual words and intermediate arguments} in \DAX s align with human judgement?''}
\FTx{To answer RQ5, % we showed a word cloud of a strongly activated filter and an n-gram selected by this filter
}
we showed n-grams extracted from input texts correctly predicted with high confidence of a strongly activated filter together with its word cloud
and asked participants to select a word (in the n-grams) fitting best with phrases in the word cloud amongst 2 words: the 
%\del{first being the word most strongly supporting the argument associated with the filter, and the second being a word attacking most strongly}
strongest supporter and attacker of the intermediate argument representing the filter.
\textbf{\textit{Results:}} \FTx{In $85\%$ \EABC{($\pm 0.7\%$)} of the answers to RQ5, participants chose the correct answer.}
%\delx{$85\%$ correct answers}
This assured us that our %notions of support and attack align 
notion of support aligns
with human judgement. %\todo{PL: Actually, we experimented with only support. Can our conclusion include attack? EA:The second word was a word that was attacking the word cloud, can we use this to support us?}
% Our experiments confirmed that in more than $96\%$ of the answers ($p << 0.001$) to RQ1 participants selected a class that the n-gram best supports in DAX, and in $85\%$ of the answers to RQ2 they choose the correct answer ($p << 0.001$).
% This assured us that our notions of support and attack agree with human perceptions.

\begin{figure}[!ht]
    \centering
    \includegraphics[width=.45\textwidth]{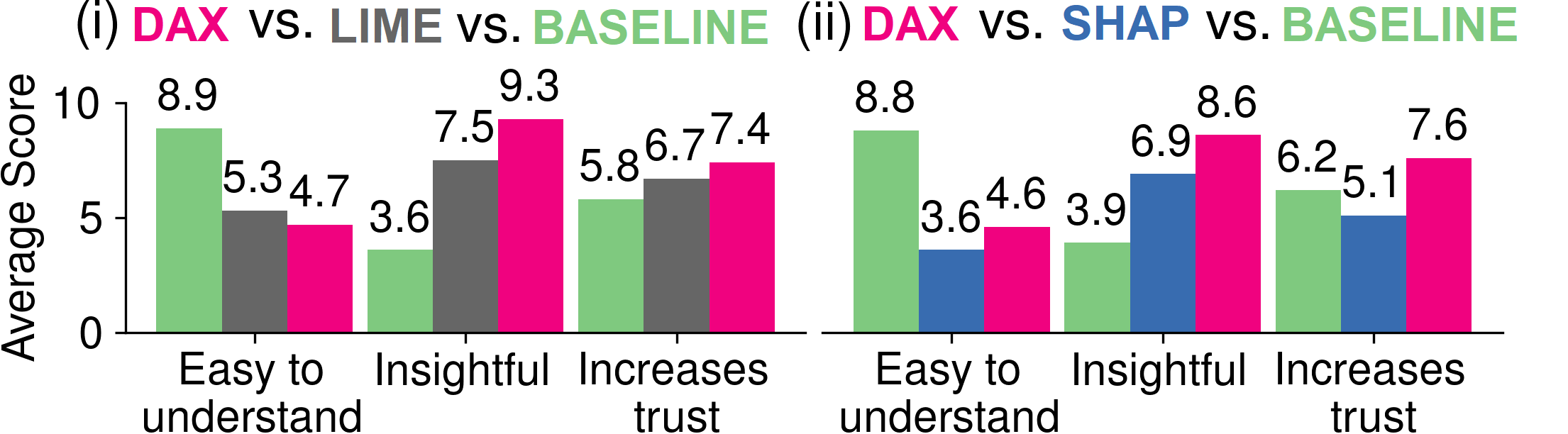}
    \caption{Participants' average scores %amongst different criteria of acceptability for an explanation 
    when comparing \DAX\ with a baseline and LIME (i) or SHAP (ii). Scores were converted from a discrete %3-value 
    scale (``No'', ``Somewhat'', ``Yes'') to numerical scores (0,5,10) for ease of comprehension.}\label{fig:experiment}
\end{figure}

% Task 5
\textbf{(C) User Acceptance.} 
To assess 
whether \emph{deep} explanations, in the form of \DAX s, are amenable to humans in comparison with ``flat'' explanations, we chose to compare \DAX s and existing methods with a somewhat similar argumentative spirit as the BAF-based \DAX s we have deployed for text classification (in particular, we chose the popular LIME and SHAP forms of explanation, presented in their standard formats within the available LIME and SHAP libraries).  To compare \DAX s and LIME/SHAP
%\delx{ how DAXs compare to existing methods} 
in terms of user acceptance, we asked participants to rate explanations according to %\delx{some} 
(desirable) criteria: 
(1) ease of understanding; 
(2) ability to provide insight into the internal %mechanism of the underlying classifier 
functioning of the CNN;
(3) capability of inspiring trust in users\nos{; and 
(4) visual appeal}.
We showed each participant an example consisting of the input text, the predicted class with its confidence, and three explanations (varying the order in different experiments): a baseline (a graph showing the probability of each class), a \DAX, and %another explanation method among LIME and SHAP.
the explanation computed by LIME or SHAP%\delx{ (chosen because they use very different styles, while being somewhat argumentative)}
. %\nos{The choice of LIME and SHAP allowed us to compare our explanation, not only with two different techniques, but also with two completely different ways of presenting the information to the user.}
% DAX scores 
\textbf{\textit{Results:}} (See Figure \ref{fig:experiment}) %According to the user \del{scores}\EA{ratings} in Figure \ref{fig:experiment}, 
As expected, \DAX s give more insight on the CNN internal working %mechanism of the classifier 
when compared with LIME ($p<0.004$) and SHAP ($p<0.006$).
\nos{DAXs are also significantly more visually appealing to users than LIME ($p<<0.001$) and SHAP ($p<0.02$).
This may be in part
due to \DAX s' use, in this instantiation, of world clouds to show filters, possible for DAXs showing the inner workings of the CNN but not for model-agnostic LIME and SHAP, focusing on the CNN's input-output behaviour only. Another possible reason for \DAX s' \FTkdd{apparent} visual appeal may be 
the interactivity of \DAX s in the instance considered, which allows %the 
users to better explore
%\delete{specific parts of the explanations more deeply to gain insight about} 
how the underlying model predicts. 
%We believe that, even if we could make LIME and SHAP interactive, they may not provide additional insight on the model due to their flat structure.
Of course, LIME and SHAP could also be made more interactive, but, given their flat structure, the interactions could only amount to incrementally showing ``attacks'' and ``supports''.}
Finally, \DAX s significantly inspire more trust %in the classifier
when compared with SHAP ($p<0.002$) although the increase in trust is not significant when compared with LIME ($p = 0.32$). %\del{We believe this is due to the background of the participants: \del{only} less than 12\% stated that they had expertise in machine learning, NNs or in explanations methods for classifiers. Therefore, most \del{of them} were probably satisfied by an easy-to-understand "flat" explanation such as LIME that provides some insight about the classifier even though it doesn't show the internal working mechanism.}
We posit that this is due to the lack of expertise of participants (less than 12\% of participants stated that they had expertise in machine learning, NNs or %explanation methods
\FTkdd{deep learning}). We also posit\FTkdd{, and leave to future studies,} %too 
that these benefits may be more significant 
%if participants were exclusively experts
with expert participants.

%%%%%%%%%%%%%%%%%%%%%%%%%%%%%%%%%%%%%%% OTHER SETTINGS $$$$$$$$$$$$$$$$$$$$$$$$$$$$$$$$$$$$$$
\iffalse
\section{Towards new \DAX\ instances}
\label{sec:guidelines}

\todo{add (spirit: DL is versitile, may hyper-parameters...so \DAX s need to.... discuss library...) SHOULD WE DISCUSS HOW WE WOULD INSTANTIATE A RNN? E.G. LEVERAGING ON PIYAWAT'S EXPERIMENTS WITH LSTMS? }

...We have considered various combinations of choices which vary the dialectical strengths, the extracted relations and resulting argumentation frameworks, satisfying different dialectical properties 
depending on the explanatory requirements of a particular application settings. ....

\todo{Properties not satisfied? shall we add something to show how this may impact the finished product with bad choices?}

\todo{Discuss why activations may not always be suitable for images or text......because they may violate dialectical properties?... ; }

\fi

%%%%%%%%%%%%%%%%%%%%%%%%%%%%%%%%%%%%%%%%%%% CONCLUSION $$$$$$$$$$$$$$$$$$$$$$$$$$$$$$$$$$$$$$$$$$$$$$$$
\section{Conclusions}
We %\delete{have} 
presented novel Deep Argumentative Explanations (\DAX s), 
%\del{a novel methodology for explanation} 
resulting from a marriage between symbolic AI (in the form of computational argumentation) and neural approaches, and leveraging on the amenability of argumentation for explanation to humans and on several advances in the %\delete{mature} 
field of computational argumentation, as well as in interpretability and visualisation of NNs.
\DAX s %are \emph{deep} explanations for NN predictions to be generated, casting 
can be used to cast
light on the inner working of NNs beyond visualisation and are adaptable to %satisfy a variety of explanatory requirements
fit a variety of explanatory settings.
We also
presented three \DAX\ instances\nos{, for: CNNs for (two forms of) 
text classification, CNNs for image classification, and FFNNs for tabular data classification% from tabular data
}%. These instances amount 
, amounting to novel stand-alone contributions per se that can be readily deployed %\todo{for explanations} for 
%shown experimentally to exhibit desirable properties in three diverse settings (text, image and tabular data) and
%%readily deployable 
in other tasks supported by the same neural architectures\nos{: future work includes exploring this further deployment}.
%
%\FT{The explanations generated by DAX (denoted DAX themselves, with an abuse of terminology) are stable~\cite{Sokol_20}  \todo{ETC?}}
We studied (existing and novel) empirical properties of our \DAX\ instances\nos{. We plan to consider further properties in future work. For example, given that our \DAX\ instances  use deterministic methods (LRP, GradCAM, activations), for the same input/output pairs they return the same information and can thus be deemed to be \emph{stable} \cite{Sokol_20}. We}, and conducted experiments, with lay participants, 
to show that \DAX s \nos{(for CNNs for text classification)  exhibit desirable properties}
can be desirable 
from the user perspective%, such as comprehensablity, usability, and increasing \AR{the} transparency \AR{of} and trust in the model
\nos{, in a stand-alone manner and in comparison with some existing explanation methods which have an argumentative spirit but are flat}. 
Future work includes exploring further \DAX\ instances, properties, and evaluation by expert users.

\nos{As future work, it would be interesting to
%Open questions for future work include 
study whether other forms of symbolic AI,
e.g., rule-based, can equally well support the  argumentative nature of explanations for NN predictions, the depth and the systematic versatility naturally supported by \DAX s.
We also plan
to conduct %further 
experiments \nos{with expert users (which may %greatly 
benefit  from \DAX s during model development) and/or} with different \DAX\ instances and tasks/data. Further, we plan to explore how \DAX s can support downstream tasks and %for debugging models
debugging, e.g., along the lines of \cite{FIND} but using the %varied spectrum of \DAX s that our framework affords
variety of explanations afforded by \DAX.  
}

\section*{Acknowledgements}
%Francesca Toni was partially supported by the Royal Academy of Engineering and by JP Morgan under the Research Chairs and Senior Research Fellowships scheme.
This research was funded in part by J.P. Morgan and  by the Royal Academy of Engineering under the Research Chairs and Senior Research Fellowships scheme.  Any views or opinions expressed herein are solely those of the authors listed, and may differ from the views and opinions expressed by J.P. Morgan or its affiliates. This material is not a product of the Research Department of J.P. Morgan Securities LLC. This material should not be construed as an individual recommendation for any particular client and is not intended as a recommendation of particular securities, financial instruments or strategies for a particular client.  This material does not constitute a solicitation or offer in any jurisdiction.  
We thank Kristijonas Cyras for helpful comments on earlier versions of this paper.

\bibliographystyle{named}
\bibliography{bib}

\newpage
\appendix 

\noindent {\bf \Huge Appendices}

\iffalse
\section{Background}

We use the LRP-0 rule as defined with a different but equivalent notation in \cite{Bach_15}. We denote with $R(j, i)$ the relevance backpropagated from neuron $j$ to neuron $i$ is $R(j, i) = \frac{a_i w_{ij}}{\sum_{l} a_l w_{lj}}$ where $a$ and $w$ denote, \resp, the activations and the weights.

SOME OF IT MAY NEED TO GO IN THE IMAGE INSTANCE HERE OR IN BODY OF PAPER, EVEN IF WE DROP THIS PART HERE??? We use the Grad-CAM weighted forward activation maps but differently from \cite{Selvaraju2016} we consider them separately for each convolutional filter and we (up)scale them to the input size, i.e., the scaled weighted forward activation map of filter $j$ wrt to the output class $o$ denoted with $G^o_j$ is defined as $G^o_j = S\left( A^j \cdot g^o_j \right)$ where $S$ denote an operator that scales to the input size, $A^j$ the activation map of filter $j$ and $g^o_j$ the Grad-CAM neuron importance weight, i.e., $g^o_j = \frac{1}{Z} \sum_x \sum_y \frac{\partial \widetilde{a}_o}{\partial A^j_{xy}}$ where $Z$ is the global average pooling coefficient and $\widetilde{a}_o$ denote the activation of class $o$ before the softmax.
\fi

\section{DAX Instances: Additional Details }
\label{app:instances}
For all models we used Keras with Adam optimizer and  set random seeds of Python (\texttt{random}), NumPy and TensorFlow/Keras to $0$.

%%%%%%%%%%%%%%%%%%%%%%% CNN TEXT $$$$$$$$$$$$$$$$$$$$$$$$
\paragraph{\bf CNN for text classification}
%As mentioned in the paper we targeted a CNN architecture with an input layer of 150 words followed by: an embeddings layer (6B GloVe embeddings of size 300); a hidden layer of 20 1D ReLU convolutional filters of which 5 are of size 2, 10 are of filter size 3 and 5 are of size 4; followed by a max pooling layer; a dropout layer with probability 0.3; and finally a dense softmax layer. 
We trained the network in batches of 128 samples for a maximum of 100 epochs with an early stopping with patience 3.
We split the IMDB dataset %\cite{imdb} 
in training (20000 samples), validation (5000) and test (25000) sets, achieving
%The CNN trained on this dataset has 
accuracy of 81.81\% on the test set.
We split the AG-News dataset %\cite{Gulli} 
in training (96000 samples), validation (24000) and test (7600) sets%. We pre-processed the AG-News dataset removing
, pre-processing to remove HTML characters, punctuation and lowering all text%. The model trained on this dataset has 
, achieving accuracy of 90.14\% on the test set\todo{.
%To generate the word clouds, we set the Pandas' random seed set to $2020$ and randomly sampled 1000 articles from the train set \FTx{(as in \cite{FIND})}.
%%%%%%
%\subsection{Hyper-parameters training} \textbf{CNN for Text Classification.}
The hyper-parameters of the text-CNNs were chosen after exploring%all the following combinations
:
%\begin{itemize}
    %\item 
    batch size $\in \!\{ 64, 128, 256, 512 \}$;
    %\item 
    patience $\in \!\{3, 5 \}$;
    %\item 
    dropout $\in \!\{ 0.0, 0.1, 0.2, 0.3,$ $ 0.5 \}$;
    %\item 
    and several filters sizes sets}\nos{$\in \{ $
    $[(10, 2), (10, 3), (10, 4)$, $(10, 5)], [(5, 2), (5, 3), (5, 4), (5, 5)]$, 
    $[(15, 2), (15, 3), (15, 4)],  
    [(20, 2),$ $ (10, 3), (5, 4)]$, 
    $[(10, 2), (10, 3), (10, 4)]$, 
    $[(12, 2), (10, 3), (8, 4)], 
    [(15, 2),$ $ (10, 3), (5, 4)], 
    [(7, 2), (7, 3), (7, 4)]$,  
    $[(5, 2), (5, 3), (5, 4)]$,  
    $[(10, 2), (10, 3), (5, 4)]$,
    $[(5, 2), (15, 3), (5, 4)]$,
    $[(5, 2), (12, 3), (5, 4)]$,
    $[(5, 2), (10, 3), (5, 4)]$,
    $[(5, 2), (10, 3), (10, 4)]$,
    $[(5, 2), (5, 3), (10, 4)]$,
    $[(15, 2), (15, 3)]$, 
    $[(20, 2), (20, 3)]$,
    $[(10, 2), (15, 3)]$, 
    $[(15, 2), (10, 3)]$,
    $[(15, 3), (10, 4)]$
    $ \}$ where $[(n_1, s_1), \ldots, (n_m, s_m)]$ represent a setting in which $n_1$ filters of size $s_1$ were used, and $n_2$ of size $s_2$, etc.}.
\paragraph{\bf CNN for Image Classification}
\label{app:CNN-image}
We used %the VGG-16 model 
VGG-16
(pre-trained on ImageNet) %\cite{imagenet_cvpr09} 
as available in %the Keras distribution we used
Keras. %The model 
This 
has a top-1 accuracy of 71.3\% and a top-5 accuracy of 90.1\% on %the ImageNet validation set \cite{KerasApplication}
ImageNet validation set.

\nos{In Step 1,
 %$\sneurons_{1}$ is such that 
 $n_{xy} \in \sneurons_1$ %is the node representing
represents the $x,y$ pixel in the input image, with $n_{xy}$  amounting to the 3 neurons corresponding to the 3 RGB channels of the pixel  ($|\sneurons_1| = 224 \cdot 224$),
 %$\sneurons_2$ is such that node 
 $n_j \in \sneurons_2$ corresponds to a filter in the last convolutional layer, where  each node is a collection of 14x14 neurons  ($|\sneurons_2| = 512$),
and  $\sneurons_{3} \!= \!\{ n_\out \}$ ($|\sneurons_3| \!=\! 1$) where $n_\out$ %is the output node, corresponding to 
is the neuron of the most probable (output) class.}

\begin{figure*}[ht]
    \centering
	%\hidebig
	{\includegraphics[width=1\textwidth]{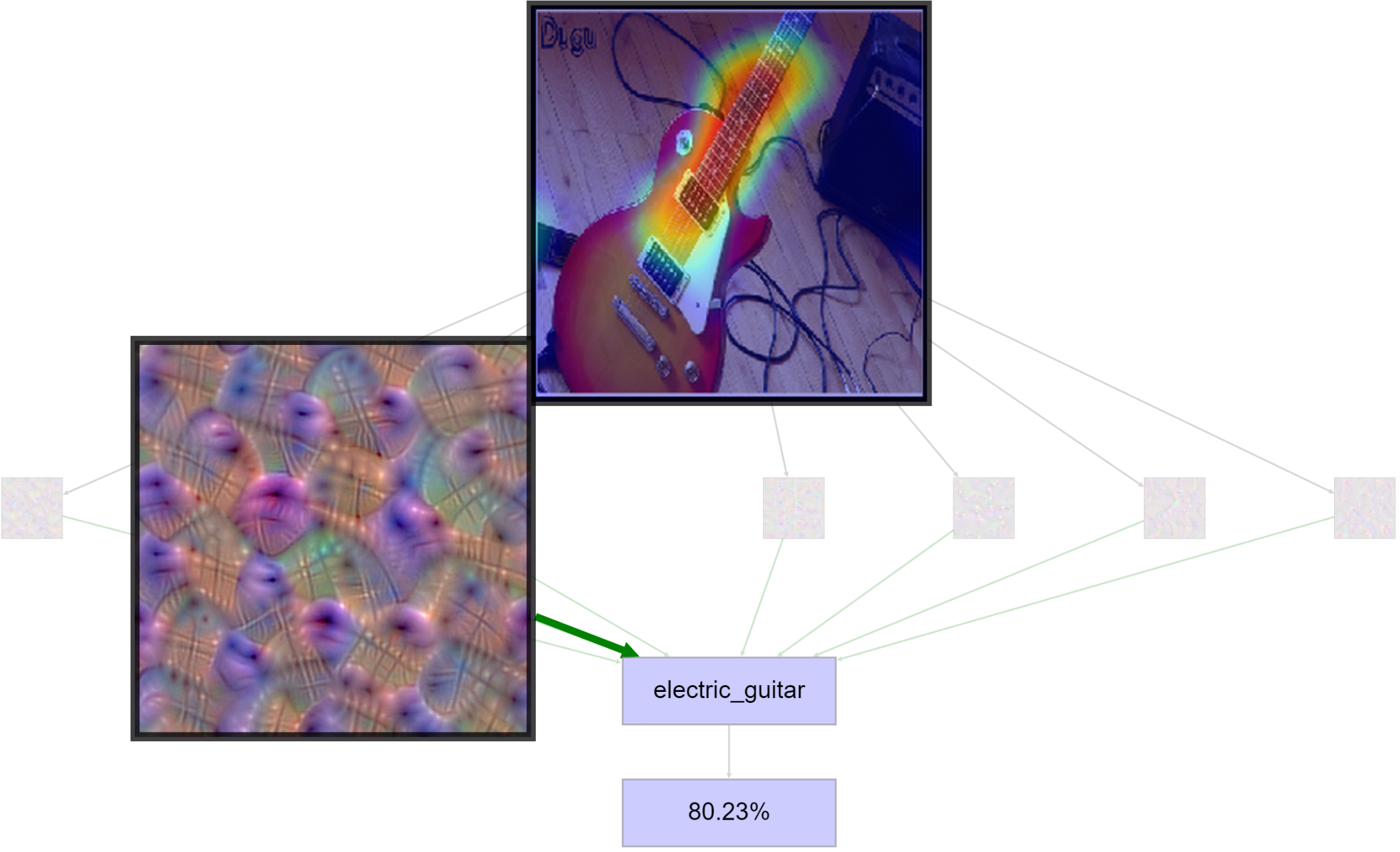}}
    \caption{
    {Graphical interactive DAX for image classification of an image from ImageNet% \textbf{after clicking on a filter activation maximization map}
    . %As in Figure \ref{fig:dax_image}, arguments are on three levels: a \emph{pixel in the input image} with a heatmap {that indicates strengths of the arguments representing that pixel influencing a convolutional filter}, the \textbf{selected} \emph{convolutional filter/activation maximization map}, and the most probable \emph{class} and its probability.  In the input image the strongest supporting pixels of the predicted class are red, then green, then blue.
    We use $\chi(\alpha)$ such that
    for argument $\alpha$ a pixel in the input image we use red for the strongest supporters (of the convolutional filter), then green, then blue (thus giving rise to a heatmap of the input image), and
    for argument $\alpha$ a filter we use \EABC{the activation maximisation of that filter}.
    The green colour of the arrows indicates support (from filters to output argument/prediction).
     }
    }
    \label{fig:dax_image2}
\end{figure*}

In \emph{Step 2}, we chose to extract SAFs $\langle \Args, \Supps \rangle$ with %a single dialectical relation of support $\Supps$.
 $\Args$  divided in 3 disjoint sets, corresponding to the strata:
$\Args_3 = \{ \arga_\out \}$, with $\arga_\out$ the \textit{output argument};
$\Args_{2}$ is the set of all \textit{intermediate arguments} $\arga_j$ representing convolutional filter $n_j$ in the the second stratum;
$\Args_{1}$ is the set of all the \textit{input arguments} $\arga_{xyj}$ representing a pixel $n_{xy}$ in the input image influencing a convolutional filter $n_j$ in the second stratum.
Overall, this gives SAFs with \FTkdd{as many as} 1,472,224 arguments \FTkdd{(as in the case of the SAF from which the \DAX\ in Figure~\ref{fig:dax_image2} is drawn)}.
We defined the relation characterisation (for %the single relation 
$\Supps$% of support in the GAF
) and the dialectical strength $\SF$ in terms of the Grad-CAM weighted forward activation maps \cite{Selvaraju2016} of the last convolutional layer, as follows. 
\EABC{Let $G^{o}_{j}$ be the (Grad-CAM) weighted forward activation map of filter $j$ (in the last convolutional layer) resized to the input size, and ${g}_j^{o}$ the Grad-CAM neuron importance weight of filter $j$ for class $c$ in the last convolutional layer.} %, and $A_k$ the feature map activations matrix of filter $k$. 
Then % relation characterisation for support ($\Rels^+$) is then defined as:
\\
\hspace*{0.5cm}
$c_{+}(n, m) = \begin{cases} 
true & \text{if } n = n_j \wedge {g}_j^{o} > 0 \\ 
true & \text{if } n = n_{xy} \wedge m = n_{j} \wedge G^{o}_{j}(x,y) > 0\\ 
false & \text{otherwise } 
\end{cases}$
%
%The dialectical strength of the arguments is defined as: 
\\
\hspace*{0.5cm}
$\SF(\arga) = \begin{cases}
\sum_{j,x,y} G^{o}_{j}(x,y) & \text{if } \arga = \arga_o \in \Args_2 \\
\sum_{x,y} G^{o}_{j}(x,y) & \text{if } \arga = \arga_j \in \Args_1\\
G^{o}_{j}(x, y) & \text{if } \arga = \arga_{xyj} \in \Args_0
\end{cases}$\\
In \emph{Step 3}, we chose %to transform the SAF into 
$\phi$ giving a graphical, interactive \DAX\ where \FTkdd{only 8 amongst the 512 intermediate arguments (%convolutional 
filters) are shown to the user (these are the 8 arguments with  highest strength)}. 
%
%Figure \ref{fig:dax_image} shows the interactive explanation of an image in its initial state in which the size of the filter activation maximization maps are proportional to the corresponding arguments strengths. 
Figure~\ref{fig:dax_image2} shows the %interactive explanation 
\DAX\ for an input image predicted by the model to be an electric guitar (with probability 80.23\%) after clicking on %the filter activation maximization map of a feature 
the left-most argument
(that seems associated with the \nos{human-comprehensible }visual concept of ``guitar cords'').
\FTkdd{We chose $\chi$ for intermediate arguments so that}
\EABC{%If $\arga_j \in \Args_\FTkdd{2}$ (i.e., when $\arga_j$ represents the convolutional filter $j$) then we used as $\chi(\arga_j)$ the activation maximization with LP-norm regularization and total variation denoising, 
\\
$\chi(\arga_j) = \argmax_{A^j} \left( \mathcal{L}_{act}(A^j) + \beta_{LP}^j \cdot \mathcal{L}_{LP}(A^j) + \beta_{LP} \cdot \mathcal{L}_{TV}(A^j) \right)$\\ where $\mathcal{L}_{act}$, $\mathcal{L}_{LP}$ and $\mathcal{L}_{TV}$ are, \resp, the activation, LP-norm and total variation losses. To compute it we used 
%the  {Keras-vis} library 
{Keras-vis} with $\beta_{LP} = 10$.
W}e ran the maximisation multiple times for filters for which the algorithm did not converge (by manual inspection) \EABC{seeding each run with the output of the previous}: we used first a jitter of 0.05 and 200 iterations, %the second a jitter 
then of 0.02 and 500 iterations, then %a jitter 
of 0.04 and 1000 iterations, then %a jitter 
of 0.05 and 1500 iterations, then %a jitter 
of 0.05 and 3000 iterations, and finally %a jitter 
of 0.01 and 1000 iterations, after which all filters' activation maximizations converged. In each run we \EABC{first set $\beta_{TV}$ to $0$ and then re-run with $\beta_{TV} = 10$ to denoise the output.}

\iffalse \begin{figure}[ht]
    \centering
	%\hidebig
	{\includegraphics[width=0.5\textwidth]{images/exampledax_image.PNG}}
    \caption{
    {Graphical interactive DAX for image classification of an image from ImageNet in its \textbf{initial state} (i.e. prior to interactions of the user with the explanation). Arguments are on three levels: \emph{pixel in the input image} with heatmap {of a pixel
    given by the sum of the (dialectical) strengths of all arguments representing that pixel}, \emph{convolutional filters/activation maximization maps} with sizes representing the corresponding arguments' strengths, and the most probable \emph{class} and its probability. The green colour of the arrows indicates support. In the input image the strongest supporting pixels of the predicted class are red, then green, then blue. }
    }
    \label{fig:dax_image}
\end{figure}
\fi

%%%%%%%%%%%%%%%%%%%%%%% FEED FORWARD $$$$$$$$$$$$$$$

\paragraph{\bf \Feedforward\ NN %for binary classification of 
with tabular data}
\label{app:FF}
%We used the  COMPAS dataset \cite{ProRepublica2016} understanding
In COMPAS, we used
\textit{sex},
\textit{age},
\textit{race},
\textit{juv\_fel\_count},
\textit{juv\_misd\_count},
\textit{juv\_other\_count},
\textit{priors\_count},
\textit{is\_recid},
\textit{is\_violent\_recid},
\textit{custody},
\textit{char\-ge\_desc},
\textit{charge\_de\-gree}
as categorical input features,
and \textit{two\_year\_re\-cid} as binary prediction.
We removed all %the 
records with unknown value and transformed into categorical variables \textit{custody}, \textit{priors\_count}, \textit{juv\_other\_co\-unt}, \textit{juv\_fel\_count} and \textit{juv\_misd\_count}.
We trained the network in batches of size 32 for a maximum of 10 epochs with early stopping with patience 5.
We split
the dataset in training (5213 samples) and test (1738 samples) sets. 
The trained (10 epochs) network has prediction accuracy of 70.2\% on the test set.
%The fact that the prediction accuracy for this dataset is relatively low %, it 
The low prediction accuracy 
\AR{did} not affect our results since %the goal of our method 
our goal \AR{was} to \emph{explain} the internal mechanism of the %classifier 
NN
in making predictions (%also of one with low accuracy
no matter how accurately). 
\todo{
The hyper-parameters of the NN were chosen after exploring the following alternatives:
%\begin{itemize}
    %\item 
    hidden layer activation function $\in \{ tanh, relu, tanh+relu \}$;
    %\item 
    batch size $\in \{ 4, 8, 16, 32, 64 \}$.
%\end{itemize}
}

In \emph{Step 1}, 
$\sneurons_1$ ($|\sneurons_1| = 58$){ with $n_f \in \sneurons_1$ corresponding to  input neuron $f$}% 
% \footnote{Here each node is the group of neurons of feature $f$}
, %a middle stratum 
$\sneurons_2$ such that $n_j \in \sneurons_2$ is the node corresponding to the hidden neuron $j$ after (tanh and) ReLU activation ($|\sneurons_2| = 8$), and %an output stratum 
$\sneurons_3 = \{ n_\out \}$ with $n_\out$ %is the output neuron of 
the most probable class node.

In {\em Step 2}, we  chose to obtain TAFs $\langle \Args, \Atts, \Supps, \Rels^{+!} \rangle$, as follows. 
%The set of arguments $\Args$ that can therefore be divided in
$\Args$  consists of 3 disjoint sets:
$\Args_3 = \{ \arga_\out \}$, where $\arga_\out$ is the \textit{output argument};
$\Args_{2}$ is the set of \textit{intermediate arguments} $\arga_{j}$ representing a %hidden neuron 
node (hidden neuron) $n_j$ in the second stratum  ;
$\Args_{1}$ is the  set of \textit{input arguments} $\arga_{ij}$ representing an input node (neuron) $n_i$ in the first stratum influencing a node %hidden neuron 
$n_j$ in the second stratum.
Let $a_{x}$ be the activation of neuron $n_x$ and $w_{xy}$ be the connection weight between %neuron 
$n_x$ and neuron $n_y$. 
%We define relation characterisations as:
Then:

$\quad\quad\quad$
$c_{-}(n_x, n_y) = 
% \begin{cases}
true \quad \text{iff } w_{xy} a_x < 0
% false & \text{otherwise}
% \end{cases}
$

$\quad\quad\quad$
$c_{+}(n_x, n_y) = 
% \begin{cases}
true \quad \text{iff } w_{xy} a_x > 0
% false & \text{otherwise}
% \end{cases}
$

$\quad\quad\quad$
$c_{+!}(n_x, n_y) = 
% \begin{cases}
true \quad \text{iff } a_y > 0 \wedge a_y - w_{xy} a_x  \leq 0
% false & \text{otherwise}
% \end{cases}
$

% Intuitively, the relation characterization is defining a critical support from $n_i$ to $n_j$ argument if deactivating $n_i$ causes the deactivation of $n_j$, this means that neuron $n_i$ is not only positively influencing $n_j$, but its contribution is essential for the activation of neuron $n_j$.

%The dialectical strength is defined as:
%\noindent and dialectical strength as:

$\SF(\arga) = \begin{cases}
a_o & \text{if } \arga = \arga_o \in \Args_2 \\
| w_{jo} \cdot a_j | & \text{if } \arga = \arga_j \in \Args_1\\
| w_{ij} \cdot w_{jo} \cdot a_i | & \text{if } \arga = \arga_{ij} \in \Args_0
\end{cases}$

\noindent Formally, 
$\langle \Args, \Atts, \Supps, \Rels^{+!} \rangle$  is \emph{%binary
counter-factuality}-compliant under %a notion of dialectical strength 
$\SF$ 
%satisfies the dialectial property of  \textit{binary counter-factuality} on a TAF $\langle \Args, \Atts, \Supps, \Rels^{+!} \rangle$ 
iff for any $\arga, \argb \in \Args$,
%such that $(\arga, \argb) \in \Rels^{+!}$ it holds that $\SF(\argb) > 0$ and $\SF(\argb) - \SF(\arga) \leq 0$.
if $(\arga, \argb) \in \Rels^{+!}$ then $\SF(\argb) \!>\! 0$ and $\SF(\argb) \!- \!\SF(\arga) \! \leq \! 0$.

\begin{figure*}[ht]
    \centering
	%\hidebig
	{\includegraphics[width=1\textwidth]{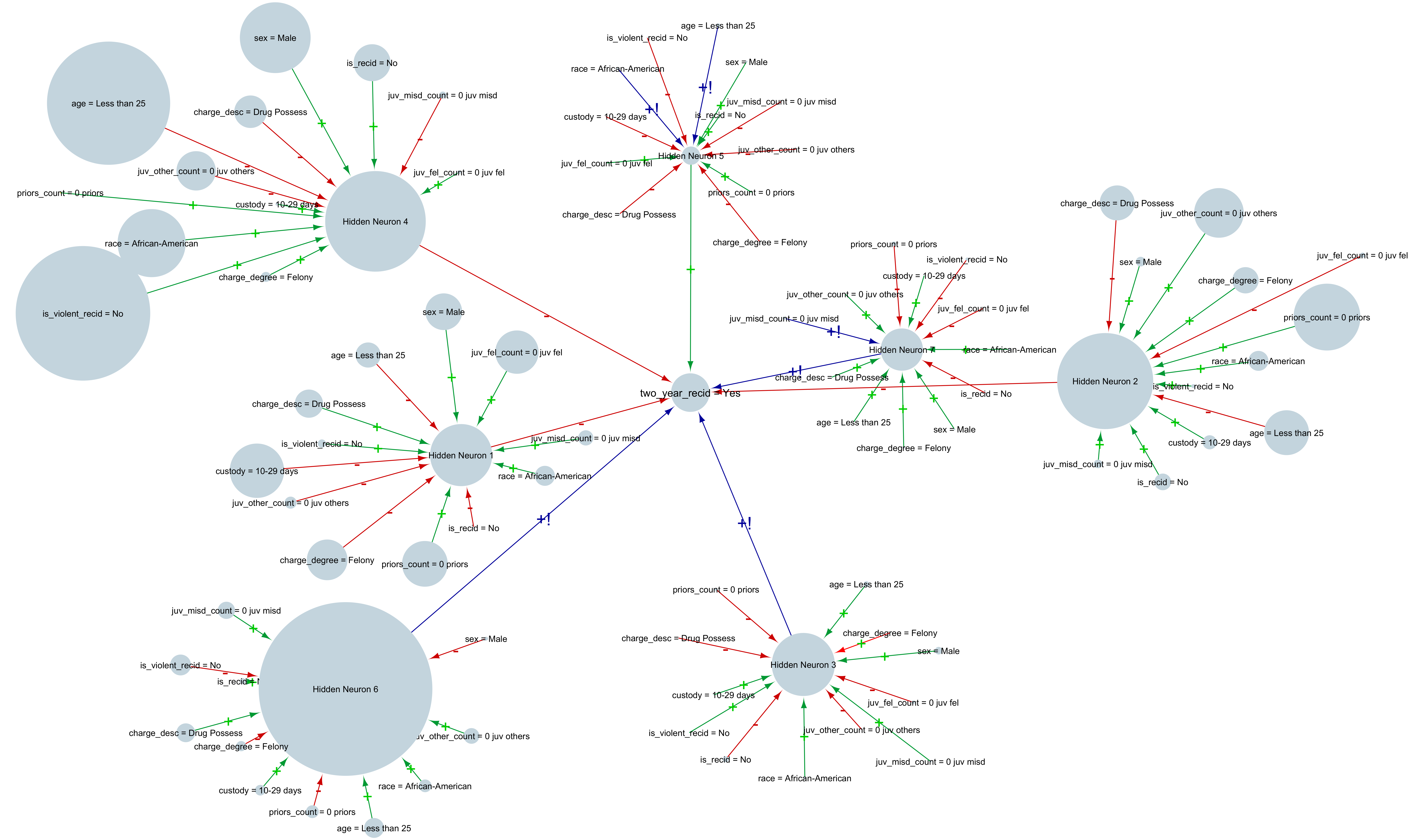}}
    \caption{%Argumentation framework from which the DAX for explaining the prediction from a sample from the COMPAS dataset is obtained. 
    A bird\AR{'s}-eye view of a TAF (with 92 arguments)
    for a  COMPAS sample.
    %The bubbles are arguments $\arga \in \Args$ with sizes corresponding to their dialectical strength $\SF(\arga)$ and as label their mapping to nodes $\rep(\arga)$ with the current value. 
    For each argument $\arga$, size corresponds to \nos{strength} $\SF(\arga)$ and label \nos{corresponds} to \nos{mapping to node} $\rep(\arga)$. Red (-), green (+) and blue (+!) arrows indicate, \resp, attack, support, and critical support.}
    \label{fig:af_ffnn}
\end{figure*}

\begin{figure*}[ht]
    \centering
	%\hidebig
	{\includegraphics[width=1\textwidth]{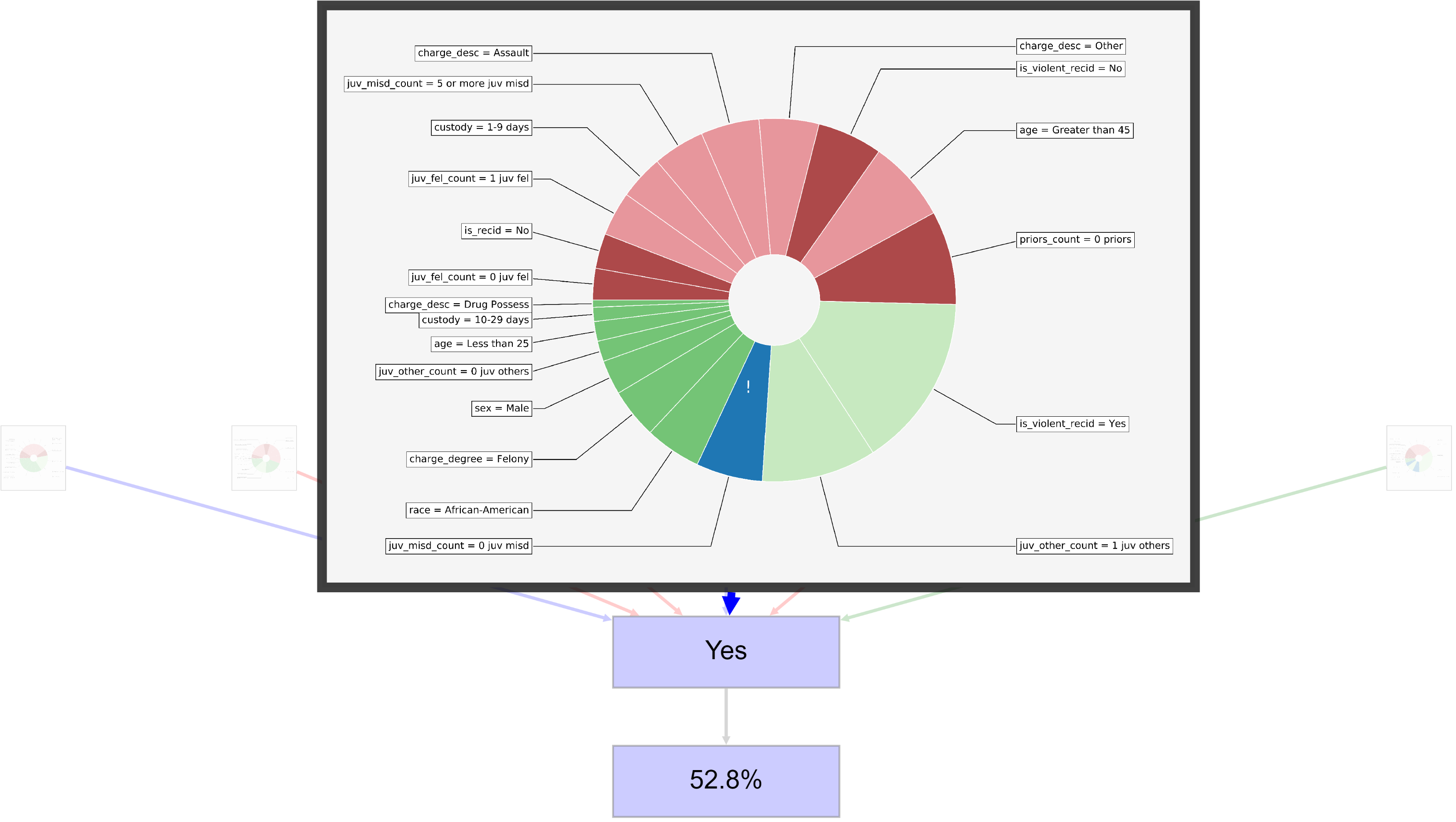}}
    \caption{
    {Graphical interactive \DAX\ %of a sample from the COMPAS test set for a FFNN for binary classification,
    for a COMPAS sample. %\textbf{after clicking on the (second from the right) pie chart}. As in Figure \ref{fig:dax_ffnn}, arguments are on three levels: the single (argument representing the) \emph{\textbf{selected} hidden neuron} (i.e. the pie chart), the arguments representing \emph{input features} visualized as the darker slices of the pie chart with size proportional to their strengths, and the most probable \emph{class} and its probability. 
    The (selected) pie chart represents, via $\chi$:   (i) an intermediate argument (hidden neuron) critically supporting the prediction 'Yes', as well as (ii) \FTkdd{the strongest attackers, supporters and critical supporters (in dark red, green and blue, \resp.) and the weakest attackers, supporters and critical supporters (in light red, green and blue, \resp) of this intermediate argument, chosen amongst the input arguments (features).}
    %In the pie chart, slices coloured of (light or dark) red represent a subset of (the most important) %(activated or non-activated) 
    %input features that minimize the activation of the hidden neuron represented by the intermediate argument, while slices coloured green or blue represent the ones that maximize it; slices coloured of dark red, dark green and blue highlight input features that for this sample are attacking, supporting and critically %/strictly 
    %supporting the intermediate argument, \resp. 
    }
    \label{fig:dax_ffnn2}}
\end{figure*}

\FTkdd{The TAFs resulting from Step 2 are clearly not human-friendly.}  In {\em Step 3}, we can %transform the TAF into a graphical explanation divided into 
generate \DAX s with 3 levels reflecting the strata structure % (see Figure~\ref{fig:dax_ffnn}): (1) the input features (represented by the darker slices of the pie charts), (2) the hidden features (represented by the pie charts), and (3) the prediction and its probability. 
of the same graphical, interactive format as for the other two instances.
%As shown in Figure \ref{fig:dax_ffnn2}, we can associate (through the mapping $\user$) a {human-comprehensible} meaning to each intermediate argument, by showing a pie chart with the input features that minimize (in red) or maximize (in green or blue) the activation of each of the hidden neurons (associated with the arguments). We represent the dialectical relations by highlighting with dark red, dark green and blue the slices of the input features (corresponding to arguments) attacking, supporting or strictly/critically supporting (\resp) the intermediate argument.
\FTkdd{As for $\user$, we can visualise arguments as shown in Figure \ref{fig:dax_ffnn2}.}
%Thus, given the peculiarity of this FFNN in which an input feature can only be 0 or 1 (because of the one-hot encoding), we joined together the visualisation of $\user$ (i.e. the meaning of intermediate arguments) and the visualisation of the input features using a pie chart (and their strength given by the size of slice).
\FTkdd{This choice deals with the peculiarity 
of this FFNN 
in which input arguments/features can only be 0 or 1 (because of the one-hot encoding) while (arguably) providing human-understandable readings of intermediate arguments/hidden neurons.  
}
\iffalse
\begin{figure*}[ht]
    \centering
	\hidebig{\includegraphics[width=\textwidth]{images/exampledax_ffnn.PNG}}
    \caption{
    Graphical interactive DAX of a sample from the COMPAS test set for a FFNN for binary classification, in its \textbf{initial state} (i.e., prior to interactions of the user with the explanation). Arguments are on three levels: those representing \emph{hidden neurons} (i.e. the pie charts) influencing the predicted class with sizes representing the corresponding their strengths, those representing \emph{input features} visualized as the darker slices of the pie charts corresponding to hidden neurons, and the most probable \emph{class} and its probability. The red, green and blue colours of the arrows indicate attack, support and critical/strict support, \resp. }
    \label{fig:dax_ffnn}
\end{figure*}
\fi

\section{Details of empirical evaluation %\delx{(Section 5)}
}
\label{app:empirical}

\FTkdd{For the  experiments in Section~\ref{sec:empirical}%\FTx{for this instance}, 
, we used 1) for the text-CNN instances, for both datasets, the test sets; 2) 
for the image-CNN, we downloaded a randomized sample of correctly predicted images from ImageNet%\delx{\footnote{The ImageNet sample we used is available in the data appendix.}}
; 3) for the FNN for COMPAS, the test set.
%\subsection{Software Platform and Computing Infrastructure}
To run all experiments, we used a Python 3.6 environment with numpy 1.18.1, pandas 1.0.3, Keras 2.2.4, TensorFlow 1.12, SpaCy 2.2.3, shap 0.31.0, Keras-vis 0.4.1, innvestigate 1.0.8 and lime 0.2.0.0 as software platform on a single Nvidia RTX 2080 Ti GPU with 11GB of memory on a machine with an Intel i9-9900X processor and 32GB of RAM.}

\textbf{Computational Costs (Image-CNN and FFNN)}.
For the CNN for image classification, given the time cost for an iteration of the feature activation maximization algorithm $c_{g}$ ($39.1 \pm 8.5$ ms, in our experiments) while we let $c$ be any other (constant or negligible) computational cost, the generation of the feature activation maximization maps (a one-time cost for $\user$) with $n_I$ iteration (we used between 200 and 3000 to generate the Figure \ref{fig:dax_image2}, depending on the filter convergence characteristics) has time complexity $O(c_{g} \cdot |\sneurons_2| \cdot n_I + c)$. Given the time cost for a prediction $c_{f}$ ($7.5 \pm 0.8$ ms, in our experiments) and that to back-propagate the output to the inputs $c_{b}$ ($7.8 \pm 0.7$ ms, in our experiments), the time complexity to generate a single \DAX\ is instead $O(c_{f} + c_{b} \cdot |\sneurons_2| + c)$.

For the FFNN for COMPAS, the generation of the base for the pie chart (a one-time cost for $\user$) has a constant time complexity since it it can be computed based on the input weights. Given the time cost for a prediction $c_{f}$ ($0.6 \pm 0.1$ ms, in our experiments) while we let $c$ be any other (constant or negligible) computational cost, the time complexity  to generate a single \DAX\ is $O(c_{f} + c)$.

%\delx{Detailed software requirements and all models and datasets we used for this paper are available in the code and data appendix.}

\section{Details of the human experiments%\delx{ (Section 6)}
}
\label{app:human}
\begin{figure}[ht]
    \centering
    \includegraphics[width=0.45\textwidth]{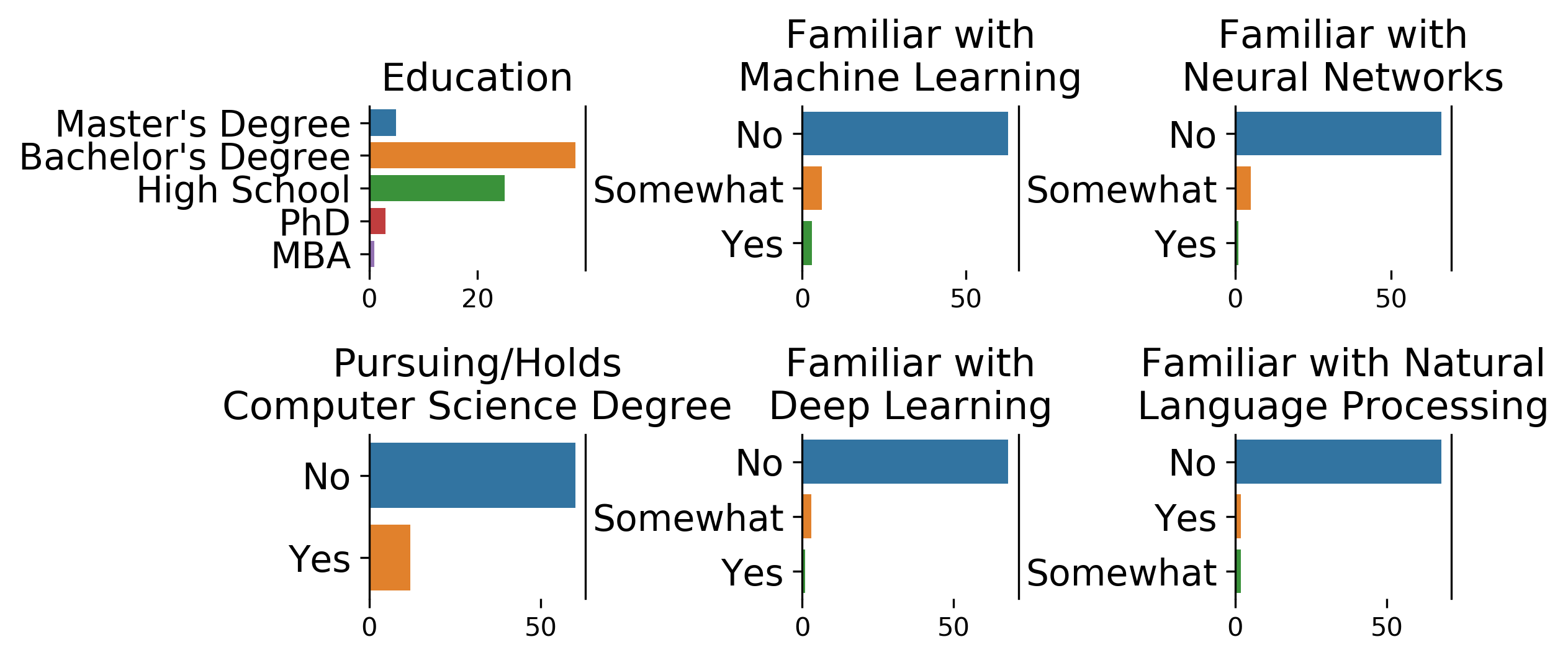}
    \caption{
	    Backgrounds of participants %in the human experiments
    }
    \label{fig:background}
\end{figure}

Figure \ref{fig:background} shows in details the background of the 72 workers of Amazon Mechanical Turk that participated in the experiments. 
% Notice that participants were filtered before any analysis to exclude low quality submissions.
\EABC{The confidence intervals and the p-values describing the significance of a score wrt another one were computed using the independent two-tailed T-test statistics.}

% \todo{Detail of the randomization of human experiments?} \EA{We already say that they are randomized}

% \todo{Link to questionnaire? } \EA{The link can potentially break the double-blind review?} \EA{The JavaScript can potentially break double blind review?}
% https://eu.qualtrics.com/jfe/form/SV_0kS1X7vQFHB13nv

% \todo{Image of the CNN architecture that we use?}

\end{document}